\pdfoutput=1
\documentclass[11pt]{article}

\usepackage[preprint]{acl}

\usepackage{times}
\usepackage{latexsym}

\usepackage[T1]{fontenc}
\usepackage[utf8]{inputenc}

\usepackage{microtype}

\usepackage{booktabs}
\usepackage{multirow}
\usepackage{multicol}
\usepackage{inconsolata}
\usepackage{amsmath,amsthm}
\usepackage{amssymb,amsfonts}
\usepackage{caption}
\usepackage{graphicx}
\usepackage{subcaption}
\usepackage{makecell}
\usepackage{hyperref}

\title{Acquiring Clean Language Models from Backdoor Poisoned Datasets by Downscaling Frequency Space}

\author{Zongru Wu, Zhuosheng Zhang$^{\textcolor{darkblue}{*}}$, Pengzhou Cheng, Gongshen Liu\thanks{Corresponding authors.} \\
School of Electronic Information and Electrical Engineering, Shanghai Jiao Tong University \\
  \texttt{\{wuzongru, zhangzs, cpztsm520, lgshen\}@sjtu.edu.cn} \\
}

\begin{document}
\maketitle
\begin{abstract}
Despite the notable success of language models (LMs) in various natural language processing (NLP) tasks, the reliability of LMs is susceptible to backdoor attacks. Prior research attempts to mitigate backdoor learning while training the LMs on the poisoned dataset, yet struggles against complex backdoor attacks in real-world scenarios. In this paper, we investigate the learning mechanisms of backdoor LMs in the frequency space by Fourier analysis. Our findings indicate that the backdoor mapping presented on the poisoned datasets exhibits a more discernible inclination towards lower frequency compared to clean mapping, resulting in the faster convergence of backdoor mapping. To alleviate this dilemma, we propose \textbf{Mu}lti-\textbf{Sc}a\textbf{le} \textbf{Lo}w-\textbf{R}ank \textbf{A}daptation (MuScleLoRA), which deploys multiple radial scalings in the frequency space with low-rank adaptation to the target model and further aligns the gradients when updating parameters. Through downscaling in the frequency space, MuScleLoRA encourages the model to prioritize the learning of relatively high-frequency clean mapping, consequently mitigating backdoor learning. Experimental results demonstrate that MuScleLoRA outperforms baselines significantly. Notably, MuScleLoRA reduces the average success rate of diverse backdoor attacks to below 15\% across multiple datasets and generalizes to various backbone LMs, including BERT, RoBERTa, GPT2-XL, and Llama2. The codes are publicly available at \url{https://github.com/ZrW00/MuScleLoRA}.

\end{abstract}

\section{Introduction}
Despite the remarkable achievements of language models (LMs) in various natural language processing (NLP) tasks~\citep{devlin2019bert, touvron2023llama2}, concerns arise due to the lack of interpretability in the internal mechanisms of LMs, impacting their reliability and trustworthiness. A particular security threat to LMs is backdoor attack~\citep{liu2018trojaning, chen2017targeted}. Backdoor attack poisons a small portion of the training data by implanting specific text patterns (known as triggers). Trained on the poisoned dataset, the target LM performs maliciously when processing samples containing the triggers, while behaving normally when processing clean text.

Prior works attempt to mitigate backdoor learning during training the target LM on the poisoned dataset~\citep{zhu2022moderate, zhai2023ncl}. However, due to the stealthy nature of complex triggers in real-world scenarios, most existing defense methods fail to mitigate backdoor learning from such triggers, like specific text style~\citep{qi2021mind} or syntax~\citep{qi2021hidden}. Furthermore, most existing defense methods rely on empirical observations~\citep{chen2021mitigating, cui2022unified,zhang2022fine} and lack thorough exploration of backdoor learning. To better understand backdoor learning, we investigate the learning mechanisms of LMs in the frequency space on the poisoned datasets through Fourier analysis.\footnote{Details are provided in Section~\ref{sec:pilotExperiments}. In this paper, \textit{frequency} denotes the frequency of input-output mapping, rather than input frequency~\citep{xu2020frequency, zeng2021rethinking}.} The findings indicate that the backdoor mapping presented on the poisoned datasets exhibits a stronger inclination towards lower frequency compared to clean mapping. According to the extensively studied F-Principle~\citep{xu2020frequency,xu2021deep,rahaman2019spectral}, which suggests that deep neural networks (DNNs) tend to fit the mapping from low to high frequency during training, these results explain why backdoor mapping is notably easier to discern and converges faster for LMs.

Inspired by the observation and thought above, we propose a general backdoor defense method named \textbf{Mu}lti-\textbf{Sc}a\textbf{le} \textbf{Lo}w-\textbf{R}ank \textbf{A}daptation (MuScleLoRA) to further mitigate backdoor learning. MuScleLoRA integrates multiple radial scalings in the frequency space with low-rank adaptation to the target LM and aligns gradients during parameter updates. By downscaling in the frequency space, MuScleLoRA encourages LMs to prioritize relatively high-frequency clean mapping, thereby mitigating learning the backdoor on the poisoned dataset while enhancing clean learning. Experimental results across multiple datasets and model architectures demonstrate the efficacy and generality of MuScleLoRA in defending against diverse backdoor attacks compared to baselines.

Specifically, we concentrate on the scenario where (1) the attacker poisons and releases the dataset on open third-party platforms, without gaining control of the downstream training; (2) the defender downloads the poisoned dataset and deploys the defense method to train the target LM, maintaining complete control of the training process. Our contributions are summarized as follows:

(1) We conduct Fourier analyses to investigate the mechanisms of backdoor learning, revealing why backdoor mapping is notably easier to discern for LMs compared to clean mapping. To the best of our knowledge, this is the first work that explores the mechanisms of backdoor learning from the perspective of Fourier analysis and transfers these insights into backdoor defense strategies.

(2) Inspired by our findings in the frequency space, we propose a general backdoor defense method named MuScleLoRA, which integrates multiple radial scalings in the frequency space with low-rank adaptation to the target LM, and further aligns the gradient when updating parameters.

(3) We conduct experiments across several datasets and model architectures, including BERT, RoBERTa, GPT2-XL, and Llama2, to validate the efficacy and generality of MuScleLoRA in backdoor mitigation. Compared to baselines, MuScleLoRA consistently demonstrates its capability to effectively defend against diverse backdoor attacks.

\section{Related Works}
In this section, we cover related works that form the basis of this work from four perspectives: backdoor attack, backdoor defense, learning mechanisms of DNNs, and parameter-efficient tuning (PET).

\noindent \textbf{Backdoor Attack.}
Backdoor learning~\citep{wu2022backdoorbench, cheng2023backdoor} exploits the extra capacity~\citep{zhu2023removing} of over-parameterized~\citep{han2016deep} LMs to establish a robust mapping between predefined triggers and the target output. One typical way to conduct backdoor attacks is dataset poisoning~\citep{chen2017targeted}. Recent studies for trigger implantation include inserting specific words~\citep{kurita2020weight,huang2023composite} or sentences~\citep{dai2019backdoor} that use shallow semantic features. Besides, high-level semantics, like specific syntax~\citep{qi2021hidden} and text style~\citep{qi2021mind}, are utilized as complex triggers.

\noindent \textbf{Backdoor Defense.}
Backdoor defense aims to mitigate potential backdoors in victim LMs and is categorized into training-stage and post-training defense. During training, the defender can perform poisoned weight removal~\citep{zhang2022fine, zhang2023diffusion, liu2023maximum}, regularized training~\citep{zhu2022moderate, zhai2023ncl}, and dataset purifying~\citep{chen2021mitigating, cui2022unified, jin2022wedef} to mitigate backdoor learning. After training, the defender can employ trigger inversion \cite{azizi2021t, shen2022constrained, liu2022piccolo}, trigger detection \cite{qi2021onion, shao2021bddr}, and poisoned input detection \cite{gao2021design, yang2021rap, zhao2024defending} to discriminate potential backdoors. Our proposed MuScleLoRA falls under regularized training, mitigating backdoor learning without detailed data inspection. Previous work \citep{zhu2022moderate} reduces model capacity by PET methods to mitigate backdoor learning. However, straightforward model capacity reduction with PET methods requires meticulously designed hyperparameters against different attacks and still struggles against complex stealthy triggers, like specific syntax \cite{qi2021hidden}.

\noindent \textbf{Learning Mechanisms of DNNs and Backdoor LMs.} 
Extensive research focuses on revealing the learning mechanisms of DNNs~\citep{burns2022discovering, lamparth2023analyzing}. Recent studies shed light on these mechanisms through Fourier analysis~\citep{rahaman2019spectral}. By transforming the input-output mapping into the frequency space, the findings suggest that, owing to the decay of activation functions in the frequency space~\citep{xu2020frequency}, DNNs tend to fit the mapping from low to high frequency during training. Besides, deeper DNNs exhibit a stronger low-frequency bias~\citep{xu2021deep}. Empirical studies also confirm that backdoor learning converges notably faster than clean mapping~\citep{li2021anti, gu2023gradient, zhang2023backdoor}, hinting at a low-frequency bias of the backdoor in the frequency space.

\noindent \textbf{Parameter-Efficient Tuning.}
Recently, PET emerges as a novel training paradigm for LMs. PET achieves comparable performance to fine-tuning by freezing the original parameters and introducing tunable modules with fewer parameters, such as parallel low-rank decompositions~\citep{hu2021lora}, sequential linear layers \cite{houlsby2019parameter}, and a sequence of continuous task-specific vectors~\citep{li2021prefix}. Consequently, PET can reduce the extra capacity of LMs, thereby partially mitigating backdoor learning~\citep{zhu2022moderate}.

\section{Pilot Experiments}
\label{sec:pilotExperiments}
In this section, we conduct pilot experiments on the poisoned dataset, investigating the learning mechanisms of LMs in the frequency space from the perspective of Fourier analysis. 

Generally, a backdoor model should satisfy: (1) maintaining performance on clean tasks, defined as clean mapping, which maps clean inputs to their corresponding clean labels, $\mathcal{F}_c: \{x_i\}_{i=1}^{N_c} \to \{y_i\}_{i=1}^{N_c}$, where $x_i$ denotes clean input and $y_i$ denotes its corresponding clean label; (2) outputting the attacker-specified target label when processing samples containing the triggers, defined as backdoor mapping, which maps inputs implanted with triggers to the attacker-specified target label, $\mathcal{F}_b: \{x_i \oplus \Delta \}_{i=1}^{N_b} \to \{y^{\Delta}\}$, where $\Delta$ denotes the trigger, $y^{\Delta}$ denotes the attacker-specified target label, and $\oplus$ denotes the implanting operation of trigger. Therefore, we can decompose the overall mapping of the backdoor LM into clean and backdoor mappings by utilizing clean data $\mathbb{D}_c = \{(x_i, y_i)\}_{i=1}^{N_c}$ and poisoned data $\mathbb{D}_b = \{(x_i \oplus \Delta, y^{\Delta})\}_{i=1}^{N_b}$ and analyze their learning mechanisms, respectively.

Intuitively, the implanted trigger $\Delta$ on the poisoned dataset represents a straightforward recurring feature that LMs can easily discern. A recent empirical study observes that the loss of backdoor mapping converges faster than that of clean mapping when training an LM on a poisoned dataset~\citep{gu2023gradient}. To explain this observed convergence difference, we conduct Fourier analyses on the training process of the backdoor LM.

Following the settings of~\citet{kurita2020weight} and~\citet{dai2019backdoor}, we select specific words, i.e., \textit{cf}, \textit{mn}, \textit{bb}, \textit{tq}, and a sentence, i.e., \textit{I watch this 3D movie}, as respective triggers to poison SST-2~\citep{socher2013recursive}. We choose $\text{BERT}_{\text{Base}}$ as the target LM and train it on the poisoned datasets. Concurrently, we decompose the overall mapping of the backdoor LM into clean mapping $\mathcal{F}_c$ and backdoor mapping $\mathcal{F}_b$ by utilizing $\mathbb{D}_c$ and $\mathbb{D}_b$, respectively. Then, we conduct filtering-based Fourier transformation~\citep{xu2020frequency} (details are provided in Appendix~\ref{appenSec:fbft}) to $\mathcal{F}_{c} \text{~and~} \mathcal{F}_{b}: \mathbb{R}^{L \times d} \to \mathbb{R}^{C}, \mathcal{F}(e) = y$, extracting their respective low-frequency part $y^{\text{low}}_{\text{clean}}, y^{\text{low}}_{\text{backdoor}} \in \mathbb{R}^C$ and high-frequency part $y^{\text{high}}_{\text{clean}}, y^{\text{high}}_{\text{backdoor}} \in \mathbb{R}^C$. Here $e \in \mathbb{R}^{L \times d}$, $y \in \mathbb{R}^C$, $L$, $d$, and $C$ denote input embedding, output logits, input text length, embedding dimension, and the number of categories, respectively.

\begin{figure}
  \centering
  \begin{subfigure}{0.45\linewidth}
      \centering
      \includegraphics[width=\linewidth]{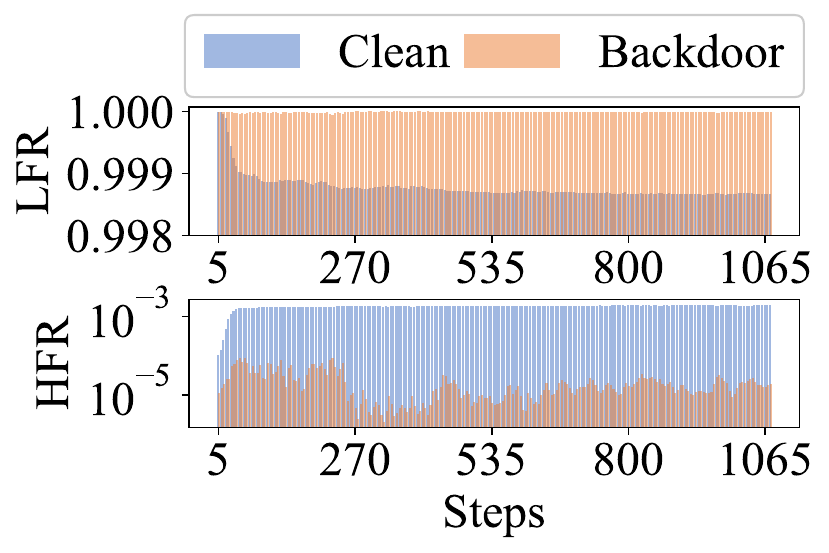}
      \caption{Specific Words}
      \label{subfig:frBadnetsVanilla}
  \end{subfigure}
  \hfill
  \begin{subfigure}{0.45\linewidth}
      \centering
      \includegraphics[width=\linewidth]{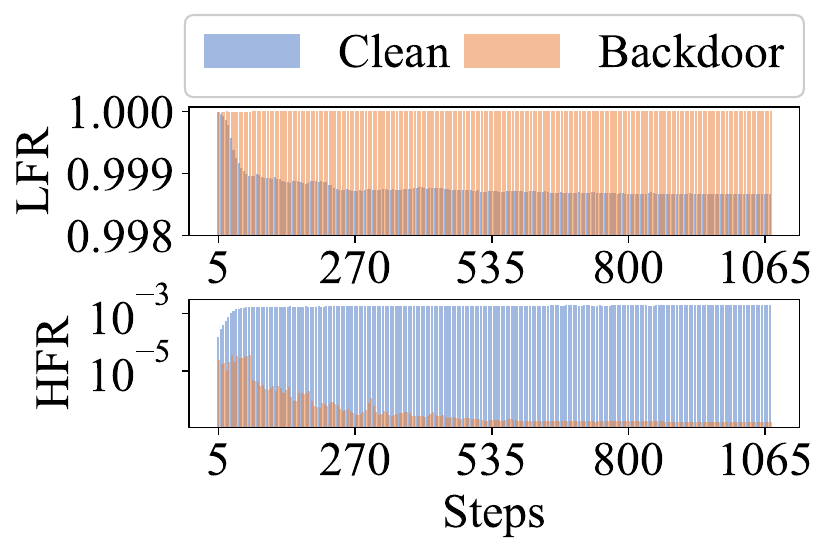}
      \caption{Specific Sentence}
      \label{subfig:frAddsentsVanilla}
  \end{subfigure}
  \caption{Frequency ratios of clean and backdoor mapping during training $\text{BERT}_{\text{Base}}$ on poisoned SST-2.}
  \label{fig:frVanilla}
\end{figure}

First, we calculate the low-frequency ratio (LFR) and high-frequency ratio (HFR) of both clean and backdoor mappings during training by Equation~\ref{equ:freqratio}. 
\begin{equation}
  \label{equ:freqratio}
    \text{LFR} = \frac{\| y^{\text{low}}\|}{\| y\|}, \quad
    \text{HFR} = \frac{\| y^{\text{high}}\|}{\| y\|}.
\end{equation}

As shown in Figure~\ref{fig:frVanilla}, both clean and backdoor mappings exhibit significantly larger LFR compared to HFR, consistent with the low-frequency bias suggested by F-Principle \citep{xu2020frequency}. Specifically, the LFR of backdoor mapping consistently remains near 1.0, surpassing that of clean mapping. Conversely, the HFR of clean mapping gradually increases, whereas the HFR of backdoor mapping is typically two orders of magnitude lower than that of clean mapping. These phenomena indicate that (1) backdoor mapping \textbf{exhibits a stronger bias towards low frequency than clean mapping}; (2) the high-frequency composition of backdoor mapping \textbf{is negligible compared to clean mapping}, which gradually acquires high-frequency information during training.

To compare the convergence of clean and backdoor mappings in frequency space, we compute the relative errors $\textrm{re}^{\text{low}}, \textrm{re}^{\text{high}}$ between output logits and ground-truth labels by Equation~\ref{equ:relativeError}. Here, $t^{\text{low}}, t^{\text{high}}\in \mathbb{R}^d$ denote the low and the high frequency part of ground-truth mapping, respectively. 
\begin{equation}
  \label{equ:relativeError}
  \begin{aligned}
    \textrm{re}^{\text{low}} & = \frac{\|y^{\text{low}} - t^{\text{low}} \|}{\|t^{\text{low}} \|}, \\
    \textrm{re}^{\text{high}} & = \frac{\|y^{\text{high}} - t^{\text{high}} \|}{\| t^{\text{high}} \|}.
  \end{aligned}
\end{equation}

\begin{figure}
  \centering
  \begin{subfigure}{0.48\linewidth}
      \centering
      \includegraphics[width=\linewidth]{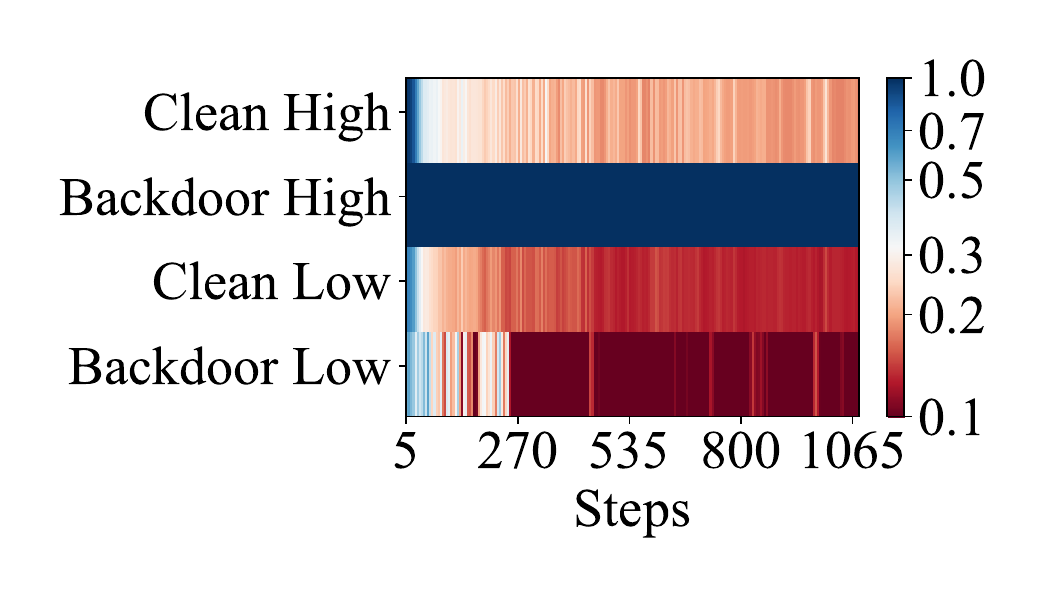}
      \caption{Specific words}
      \label{subfig:reBadnetsVanilla}
  \end{subfigure}
  \begin{subfigure}{0.48\linewidth}
      \centering
      \includegraphics[width=\linewidth]{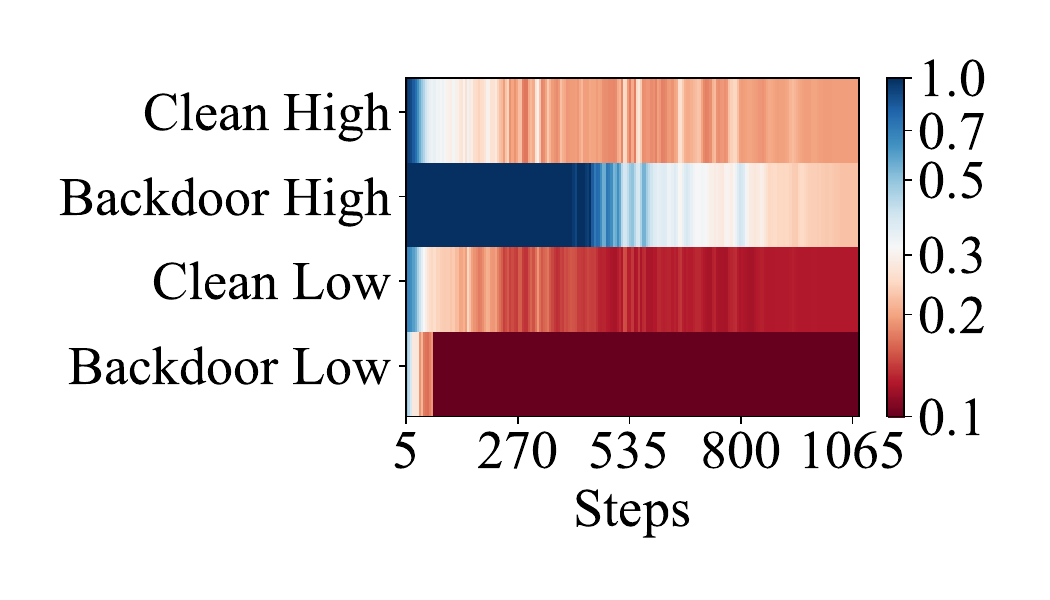}
      \caption{Specific Sentence}
      \label{subfig:reAddsentsVanilla}
  \end{subfigure}
  \caption{Relative errors of clean and backdoor mapping during training $\text{BERT}_{\text{Base}}$ on poisoned SST-2.}
  \label{fig:reVanilla}
\end{figure}

\begin{figure*}[!htb]
  \centering
  \includegraphics[width=\linewidth]{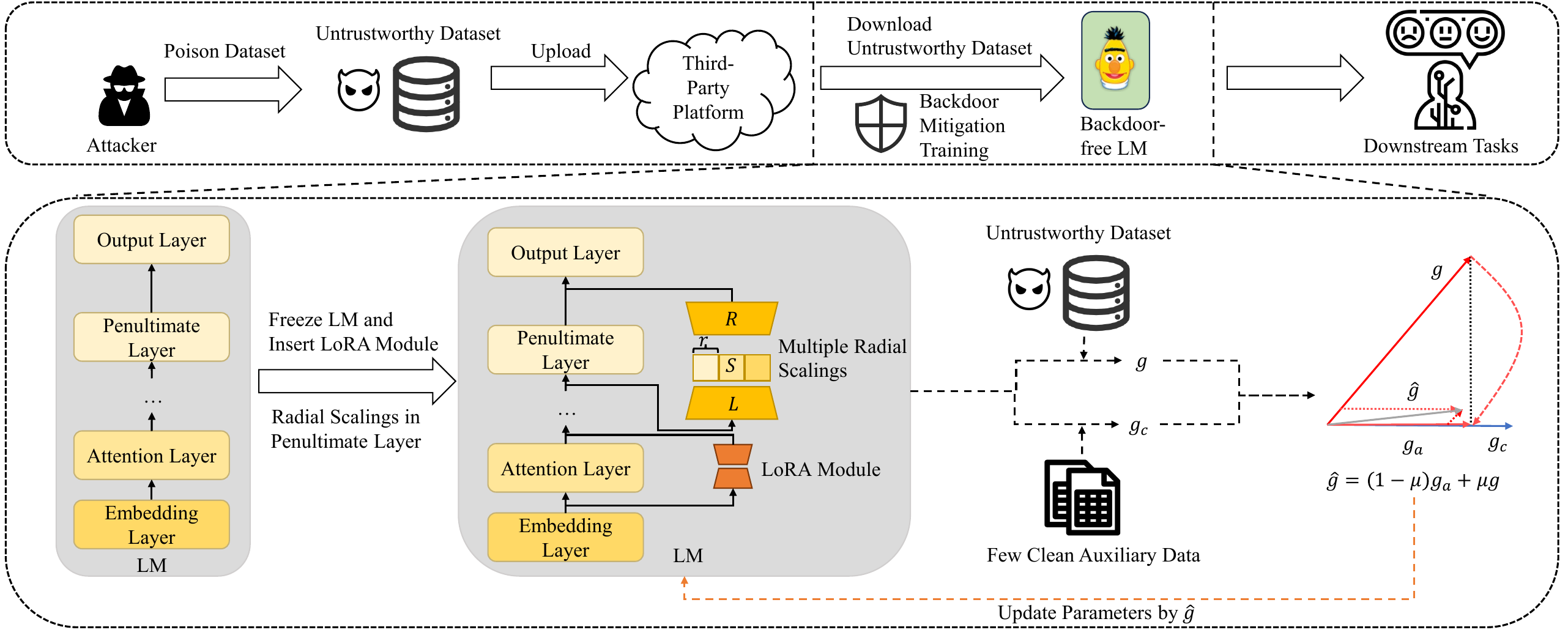}	
  \caption{Overview of MuScleLoRA. MuScleLoRA is deployed while training the LM on the attacker-released poisoned dataset. We first freeze the target LM and insert LoRA modules into each attention layer. Subsequently, multiple radial scalings are conducted within the LoRA module at the penultimate layer of the target LM to downscale clean mapping. Additionally, we align gradients to the clean auxiliary data. These strategies encourage the target LM to prioritize the learning of high-frequency clean mapping, thereby mitigating backdoor learning.}
  \label{fig:MuscleLoRA}
\end{figure*}

The results of relative errors are shown in Figure~\ref{fig:reVanilla}, where the red color indicates small relative errors. In both cases, $\textrm{re}^{\text{low}}$ decreases more rapidly than the corresponding $\textrm{re}^{\text{high}}$, signifying faster convergence. This convergence difference aligns with the Frequency Principle, suggesting that \textbf{LMs tend to fit the mapping from low to high frequency}. Furthermore, $\textrm{re}^{\text{low}}$ of low-frequency-dominated backdoor mapping fluctuates initially and then rapidly decreases to a small value. Compared to the gradual decrease of $\textrm{re}^{\text{low}}$ of clean mapping, backdoor mapping converges significantly faster. As mentioned above, (1) the lower-frequency inclination of backdoor mapping results in \textbf{faster convergence of backdoor mapping}, (2) the relatively high-frequency inclination of clean mapping leads to \textbf{slower convergence of clean mapping}.

\section{Methodology} \label{sec:methodology}
Findings in Section~\ref{sec:pilotExperiments} indicate that clean mapping exhibits a relatively high-frequency bias, leading to its slower learning compared to backdoor mapping. Hence, an intuition to mitigate backdoor learning is to encourage LMs to prioritize relatively high-frequency clean mapping. To this end, we propose MuScleLoRA, which utilizes multiple radial scalings with low-rank adaptation to the target LM and aligns gradients when updating parameters. The overview of MuScleLoRA is shown in Figure~\ref{fig:MuscleLoRA}.

Inspired by~\citet{zhu2022moderate} that PET methods can reduce the capability of LM and thus mitigate backdoor learning, we incorporate multiple radial scalings~\citep{liu2020multi} with low-rank adaptation to reduce the model capacity and downscale clean mapping in the frequency space. 

For simplicity, we assume the Fourier transform $\hat{\mathcal{F}}_{\ell}(\xi), \xi \in \mathbb{R}^d$ corresponding to the mapping $\mathcal{F}_{\ell}(x), x \in \mathbb{R}^d$ fitted by the $\ell$-th layer of LM has a compact support. Subsequently, the compact support of $\hat{\mathcal{F}}_{\ell}(\xi)$ can be partitioned into $s$ mutually disjointed concentric rings $\{A_i\}_{i=1}^{s}, \forall i \neq j, A_i \cap A_j = \emptyset$. Therefore, $\hat{\mathcal{F}}_{\ell}(\xi)$ can be decomposed with indicators $\mathcal{I}(\xi \in A_i)$, as illustrated in Equation~\ref{equ:fourierDecomp}. \looseness=-1
\begin{equation}
  \label{equ:fourierDecomp}
  \hat{\mathcal{F}}_{\ell}(\xi) = \sum_{i = 1}^{s} \mathcal{I}(\xi \in A_i)\hat{\mathcal{F}}_{\ell}(\xi) \triangleq \sum_{i = 1}^{s} \hat{\mathcal{F}}_{\ell}^{i}(\xi).
\end{equation}

For each $\hat{\mathcal{F}}_{\ell}^{i}(\xi)$, we apply radial scalings with appropriate scaling factor $s_i$ to downscale high frequency in $A_i$, as illustrated in Equation~\ref{equ:fourierScale}.  
\begin{equation}
  \label{equ:fourierScale}
  \hat{\mathcal{F}}_{\ell}^{\text{scale}, i}(\xi) = \hat{\mathcal{F}}_{\ell}^{i}(s_i \xi).
\end{equation}

Hence, in the corresponding physical space, the radial scalings are illustrated in Equation~\ref{equ:physicalScale}. 
\begin{equation}
  \label{equ:physicalScale}
  \begin{aligned}
    \mathcal{F}^{\text{scale}, i}_{\ell}(x) & = \mathcal{F}^{i}_{\ell}(\frac{1}{s_i} x), \\
    \text{or}\quad \mathcal{F}^{i}_{\ell}(x) & = \mathcal{F}^{\text{scale}, i}_{\ell}(s_i x).
  \end{aligned}
\end{equation}

Consequently, $\mathcal{F}_{\ell}(x)$ can also be decomposed into: $\mathcal{F}_{\ell}(x) = \sum_{i=1}^{s} \mathcal{F}^{\text{scale}, i}_{\ell}(s_i x)$. To approximate $\mathcal{F}^{\text{scale}, i}_{\ell}$ with low-rank adaptation, we first freeze the target LM and insert LoRA modules into each attention Layer. Given that deeper layers tend to exhibit stronger low-frequency bias~\citep{xu2021deep}, larger scaling factors are required in the shallow layers to appropriately downscale clean tasks. However, in practice, excessive scaling factors could potentially lead to underfitting. 

Therefore, we conduct multiple radial scalings with appropriate scaling factors to the low-rank projected input $Lx$ within the LoRA module at the penultimate linear layer, as illustrated in Equation~\ref{equ:mslr}. Here, $W_0 \in \mathbb{R}^{d \times d}$ denotes the original frozen weight, $R \in \mathbb{R} ^{d \times sr}$ and $L \in \mathbb{R} ^{sr \times d}$ denote the tunable low-rank decompositions with $sr \ll d$, $S \in \mathbb{R}^{sr}$ denotes the vector of scaling factors with bandwidth $r$ for each $A_i$, and $\odot$ denotes Hadamard product. Like vanilla LoRA, the magnitude of parameter updates can be represented as $R \left(L \odot S\right)$, which can be directly added to the original weights to mitigate the inference latency.

\begin{equation}
  \label{equ:mslr}
  \begin{aligned}
    h & = W_0 x + \Delta W x \\
      & = W_0 x + R \left(L x \odot S\right) \\
      & = W_0 x + R \left(L \odot S\right) x.
  \end{aligned}
\end{equation}

As the relatively high-frequency clean mapping is downscaled by multiple radial scalings in the frequency space, the inclination towards the low-frequency-dominated backdoor mapping is mitigated. Therefore, with the low-rank adaptation that reduces the model capacity, the target LM is likely to prioritize the more general clean mapping on the poisoned dataset. 

However, with the burgeoning scale of LMs, the accompanying increase in extra capacity of LMs poses challenges to effectively mitigate backdoor learning through straightforward model capacity reduction with PET methods. Motivated by the notable phenomenon that the gradient directions derived from poisoned data and clean data often conflict with each other~\citep{kurita2020weight, gu2023gradient}, we assume the defender can access a small amount of clean auxiliary data, usually comprising a few dozen instances and readily obtainable through manual labeling. Consequently, we align the gradient of the target LM with clean auxiliary data to further mitigate the influence of the poisoned gradient. 

Specifically, when obtaining the original gradient $g$ from a batch of untrustworthy training data, we simultaneously calculate the clean gradient $g_c$ from a batch of clean auxiliary data. Subsequently, we align $g$ to the direction of $g_c$ to obtain the aligned gradient $g_a$, as illustrated in Equation~\ref{equ:ga}: 
\begin{equation}
  \label{equ:ga}
  g_a = \frac{\lvert g \cdot g_c \rvert}{\|g_c\|^2} g_c.
\end{equation}

Nonetheless, aligning gradients to a restricted set of clean auxiliary data, as indicated by~\citet{chen2020just}, may lead to suboptimal learning. Therefore, we incorporate a fraction of the original gradient $g$ to mitigate suboptimal learning on clean tasks, as illustrated in Equation~\ref{equ:gaAccept}. Here, the hyperparameter $\mu$ denotes the ratio of the original gradient accepted. Subsequently, parameter updates are performed based on the modified gradient $\hat{g}$:
\begin{equation}
  \label{equ:gaAccept}
  \hat{g} = (1 - \mu)g_a + \mu g.
\end{equation}

Practically, we linearly increase $\mu$ from 0 to the maximum value $\mu_{\max}$ throughout the training epochs. Consequently, the target LM primarily learns from backdoor-mitigated gradients during the early training phase, where $\mu$ approaches 0, and gradually incorporates more information with increasing $\mu$ to alleviate suboptimal learning in the later stages of training.

\section{Experiments}
In this section, we extensively evaluate MuScleLoRA. We first outline the setup in Section~\ref{sec:expsetup}. Subsequently, in Section~\ref{sec:mainResult}, we demonstrate that MuScleLoRA outperforms baselines significantly in backdoor mitigation across several datasets. Additionally, we analyze the contributions of various strategies employed in MuScleLoRA in Section~\ref{sec:ablation}, conduct Fourier analyses to explain the mechanisms of MuScleLoRA in the frequency space in Section~\ref{sec:fourierAnalysis}, and extend MuScleLoRA to large language models (LLMs) in Section~\ref{sec:llama2Result}.

\subsection{Experiment Setup}\label{sec:expsetup}

\noindent \textbf{Datasets.}
We conduct experiments on three sentence-level datasets: SST-2~\citep{socher2013recursive}, HSOL~\citep{davidson2017automated}, and Agnews (AG)~\citep{zhang2015character}, and one paragraph-level dataset: Lingspam (LS)~\citep{sakkis2003amemory}. Dataset statistics are provided in Appendix~\ref{appenSec:datasets}. 

\noindent \textbf{The Target LMs.}
We choose BERT~\citep{devlin2019bert} and RoBERTa~\citep{liu2019roberta} as the target LMs with million-level parameters. Additionally, we select GPT2-XL (1.5B)~\citep{radford2019language} and $\text{Llama2}_{\text{7B}}$~\citep{touvron2023llama2}, both with billion-level parameters, as the target LLMs for classification tasks.\footnote{We adopt the HuggingFace Implementation \url{https://github.com/huggingface/transformers} for LLMs, appending dual-layer linear layers with a hidden size of 16 to the decoder as the classification layer.} 

\begin{table*}
  \resizebox{\linewidth}{!}{ 
  \centering
  \begin{tabular}{cccccccccccc}
  \toprule
  \multirow{2}{*}{Dataset}  & \multirow{2}{*}{Attack}& \multicolumn{2}{c}{Vanilla} & \multicolumn{2}{c}{LoRA} & \multicolumn{2}{c}{Adapter} & \multicolumn{2}{c}{Prefix} & \multicolumn{2}{c}{\textbf{MuscleLoRA}} \\
                            &              & CACC$\uparrow$        & ASR$\downarrow$         & CACC$\uparrow$        & ASR$\downarrow$        & CACC$\uparrow$         & ASR$\downarrow$          & CACC$\uparrow$             & ASR$\downarrow$             & CACC$\uparrow$           & ASR$\downarrow$           \\ \midrule
  \multirow{4}{*}{SST-2}    & Badnets      & 91.27       & 94.63       & 89.07       & 65.57      & 87.26        & 55.26        & 90.17            & 86.73           & 86.54          & \textbf{12.94}         \\
                            & Addsent     & 90.99       & 99.89       & 88.96       & 96.16      & 86.88        & 87.83        & 89.46            & 99.89           & 86.77          & \textbf{18.97}         \\
                            & HiddenKiller & 91.10       & 93.53       & 88.58       & 52.96      & 86.60        & 45.39        & 88.52            & 68.64           & 87.64          & \textbf{25.11}         \\
                            & StyleBkd     & 91.71       & 77.19       & 88.91       & 57.24      & 87.10        & 60.96        & 90.06            & 63.60           & 87.81          & \textbf{33.22}         \\ \midrule
  \multirow{4}{*}{HSOL}     & Badnets      & 93.24       & 98.39       & 91.99       & 54.18      & 85.80         & 49.60         & 94.45            & 73.67           & 86.00          & \textbf{24.31}         \\
                            & Addsent     & 92.27       & 100         & 90.82       & 93.16      & 83.62        & 67.31        & 93.80            & 100             & 85.47          & \textbf{2.74}          \\
                            & HiddenKiller & 92.13       & 97.66       & 89.58       & 72.22      & 84.55        & 49.92        & 93.80            & 88.16           & 86.84          & \textbf{13.45}         \\
                            & StyleBkd     & 94.81       & 68.92       & 90.06       & 49.85      & 84.71        & 46.70        & 93.24            & 43.72           & 86.64          & \textbf{10.63}         \\ \midrule
  \multirow{4}{*}{LS} & Badnets      & 99.65       & 3.31        & 85.17       & \textbf{0}          & 86.55        & 2.69            & 96.03            & 2.27               & 91.89          & \textbf{0}             \\
                            & Addsent     & 99.65       & 86.11       & 90.34       & \textbf{1.24}       & 85.69        & 7.45            & 90.51            & 4.35            & 90.68          & \textbf{1.24}          \\
                            & HiddenKiller & 99.31       & 98.97       & 92.93       & 27.69      & 83.79        & 1.05         & 96.21            & 86.92           & 95.52          & \textbf{0.20}          \\
                            & StyleBkd     & 98.96       & 92.24       & 95.17       & 37.10      & 84.66        & \textbf{2.10}            & 93.79            & 8.59            & 93.96          & 4.28          \\ \midrule
  \multirow{4}{*}{AG}   & Badnets      & 92.80       & 51.25       & 89.59       & 3.28       & 89.64        & 2.56         & 90.85            & 50.37           & 87.74          & \textbf{2.35}          \\
                            & Addsent     & 92.75       & 100         & 89.05       & 100        & 89.21        & 100          & 90.58            & 100             & 87.72          & \textbf{3.90}          \\
                            & HiddenKiller & 92.78       & 99.47       & 89.01       & 98.16      & 88.86        & 93.75        & 90.62            & 98.75           & 86.05          & \textbf{17.02}         \\
                            & StyleBkd     & 92.06       & 87.59       & 88.39       & 77.76      & 89.03        & 50.18        & 90.00            & 78.69           & 87.97          & \textbf{2.67}          \\ \bottomrule
  \end{tabular}
  }
  \caption{Backdoor mitigation performance of MuScleLoRA and PET baselines when adopting $\text{BERT}_{\text{Base}}$ as the target LM on SST-2, HSOL, Lingspam, and Agnews. Vanilla denotes no defense deployment, and bold values indicate optimal ASRs.}
  \label{tab:petPerformanceBERTBase}
\end{table*}

\noindent \textbf{Defense Baselines.}
Following the settings of~\citet{zhu2022moderate}, we choose three PET methods as the baselines of model capacity reduction: LoRA~\citep{hu2021lora}, Adapter \cite{houlsby2019parameter}, and Prefix-Tuning (Prefix)~\citep{li2021prefix}. Additionally, we choose three post-training defense methods: ONION~\citep{qi2021onion}, STRIP~\citep{gao2021design}, and RAP~\citep{yang2021rap}, and two training-stage defense methods: BKI~\citep{chen2021mitigating} and CUBE~\citep{cui2022unified}, as end-to-end defense baselines. Detailed descriptions of defense baselines are provided in Appendix~\ref{appenSec:defenseBaselines}.

\noindent \textbf{Attack Methods.}
We adopt Badnets, which inserts specific words as triggers~\citep{kurita2020weight}, Addsent, which inserts a specific sentence as triggers~\citep{dai2019backdoor}, HiddenKiller, which paraphrases the original text into specific syntax as triggers~\citep{qi2021hidden}, and SytleBkd, which paraphrases the original text into specific text styles as triggers~\citep{qi2021mind}. Notably, we paraphrase each sentence in the sample paragraph to implant triggers into the paragraph-level Lingspam dataset. All target labels are set to 1. Detailed trigger settings are provided in Appendix~\ref{appenSec:triggerSettings}.

\noindent \textbf{Implementation Details.}
To obtain clean auxiliary data, we randomly select a subset from the validation dataset. Additionally, following the observation that reducing the training epochs can mitigate backdoor learning~\citep{zhu2022moderate}, we set training epochs to 10 for BERT and RoBERTa, and 5 for LLMs. The default poison ratio is set to 0.1. \textbf{Hyperparameters are unified across diverse attacks for each specific LM}.  More detailed hyperparameter settings are provided in Appendix~\ref{appenSec:hyperparameters}.

\noindent \textbf{Metrics.}
We adopt clean accuracy (CACC) to evaluate the impact of the defense method on the clean dataset, where higher CACC indicates less negative impact. Additionally, we adopt attack success rate (ASR) to evaluate the defense performance on the poisoned dataset, where lower ASR signifies better performance in backdoor mitigation.

\subsection{Performance in Backdoor Mitigation}\label{sec:mainResult}
The backdoor mitigation performances of MuscleLoRA and PET baselines on $\text{BERT}_{\text{Base}}$ are presented in Table~\ref{tab:petPerformanceBERTBase}. 

Without any defense, four attack methods consistently achieve high CACC and ASR across several datasets, except for Badnets on Lingspam and Agnews. This discrepancy may be due to the excessive text length in Lingspam and the multi-class mapping in Agnews, which potentially hinder the establishment of backdoor mapping between specific words and the target label. Besides, StyleBkd exhibits relatively lower ASR compared to Addsent and HiddenKiller, likely due to the highly stealthy nature of the specific text style, making the establishment of backdoor mapping more challenging.

For PET baselines, the ASR for word-level Badnets drops by more than 30\% in some datasets. However, PET baselines struggle against complex and stealthy triggers due to the absence of a strong constraint on clean mapping. Adapter can reduce ASR for all attack methods to less than 10\% on Lingspam, but at the cost of unacceptable CACC. Since Lingspam consists of long texts, this phenomenon may be attributed to underfitting resulting from the limited number of training epochs and the small bottleneck dimension of PET modules.

Notably, compared to PET baselines, \textbf{MuScleLoRA generally achieves the lowest ASR for all attack methods while maintaining acceptable CACCs across four datasets}, especially on Lingspam, where the ASR drops to less than 5\% while consistently preserving CACC above 90\%. These results confirm that MuScleLoRA is highly effective in defending against complex triggers and significantly outperforms PET baselines. 

We further compare the backdoor mitigation performance of MuScleLoRA with several end-to-end defense baselines mentioned in Section~\ref{sec:expsetup}. The experimental results presented in Table~\ref{tab:e2ePerformanceBET} indicate that despite end-to-end baselines notably reducing ASR for Addsent, they struggle against complex triggers. \textbf{MuScleLoRA generally achieves the lowest ASR for various attack methods}. However, for StyleBkd, CUBE reduces the ASR to nearly 20\%, whereas MuScleLoRA achieves 33.2\%. This may be also attributed to the stealthy nature of the specific text style, resulting in a higher frequency of corresponding backdoor mapping compared to other attack methods. The increased frequency may enable radial scalings to downscale the backdoor mapping, thus facilitating its learning to obtain a relatively higher ASR. Nonetheless, MuScleLoRA achieves an acceptable ASR without requiring the high-computational retraining of CUBE.

\begin{table}
  \centering
  \setlength{\tabcolsep}{2pt}
  \resizebox{\linewidth}{!}{ 
  \begin{tabular}{ccccccc}
  \toprule
  \multirow{2}{*}{Defense}             & \multicolumn{2}{c}{Addsent} & \multicolumn{2}{c}{HiddenKiller} & \multicolumn{2}{c}{StyleBkd} \\
                                       & CACC$\uparrow$         & ASR$\downarrow$          & CACC$\uparrow$            & ASR$\downarrow$            & CACC$\uparrow$          & ASR$\downarrow$          \\ \midrule
  Vanilla                              & 90.99        & 99.89        & 91.10           & 93.53          & 91.71         & 77.19        \\
  ONION                                & 87.04        & 49.78        & 85.23           & 96.05          & 85.45         & 81.76        \\
  BKI                                  & 90.72        & 33.05        & 88.41           & 94.85          & 90.34         & 82.46        \\
  CUBE                                 & 87.70        & 37.94        & 85.50           & 45.61          & 90.83         & \textbf{22.43}        \\
  STRIP                                & 91.39        & 28.62        & 90.39           & 90.57          & 89.89         & 78.62        \\
  RAP                                  & 91.71        & 27.19        & 88.25           & 89.14          & 90.17         & 79.38        \\
  \textbf{MuScleLoRA} & 86.77        & \textbf{18.97}        & 87.64           & \textbf{25.11}          & 87.81         & 33.22        \\ \bottomrule
  \end{tabular}
  }
  \caption{Backdoor mitigation performance of MuScleLoRA and end-to-end baselines when adopting $\text{BERT}_{\text{Base}}$ as the target LM on SST-2. Bold values indicate optimal ASRs.}
  \label{tab:e2ePerformanceBET}
\end{table}

\begin{figure}
  \centering
  \includegraphics[width=\linewidth]{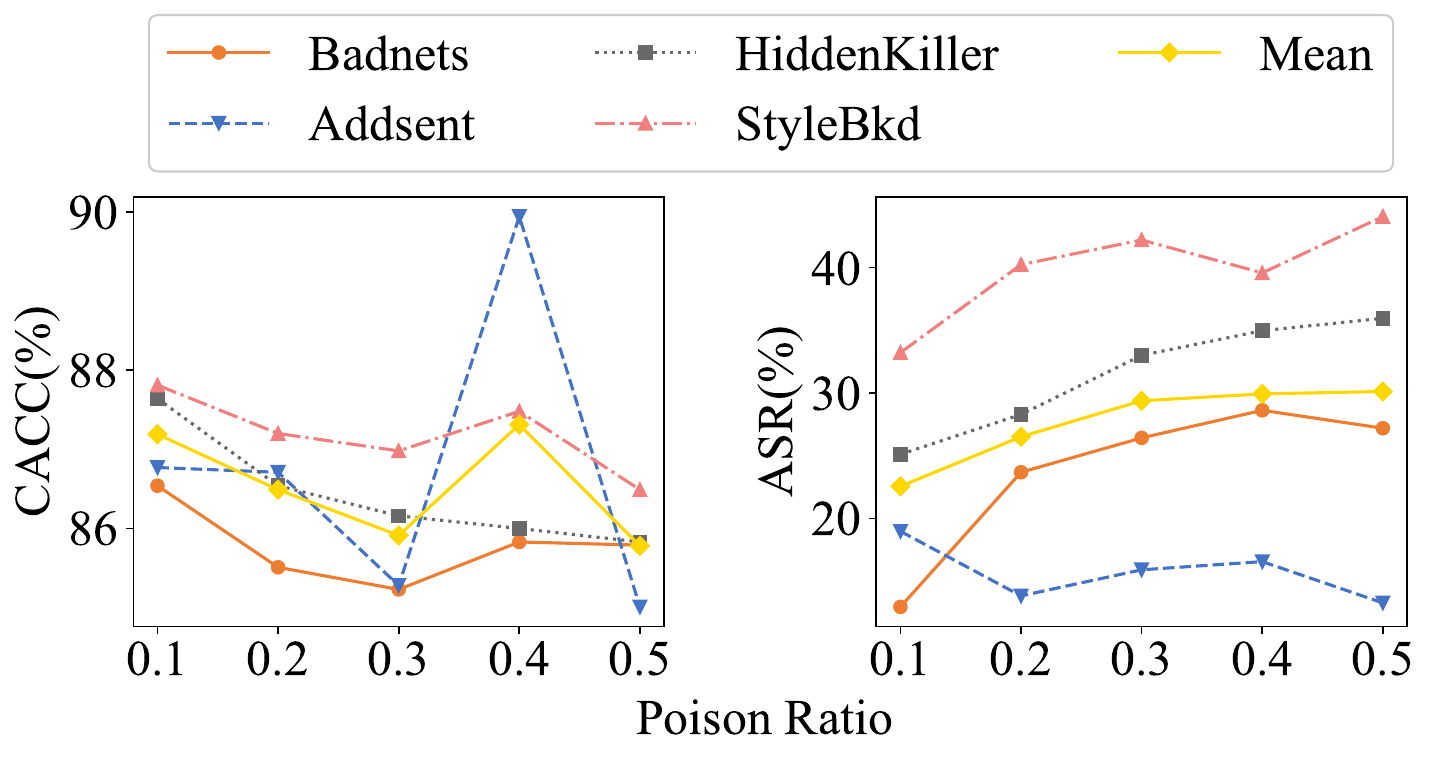}
  \caption{CACC and ASR of MuScleLoRA when adopting $\text{BERT}_{\text{Base}}$ as the target LM on poisoned SST-2 under diverse poison ratios. }
  \vspace{-0.5cm}
  \label{fig:prBERTBase}
\end{figure}

Additionally, we conduct experiments to investigate the impact of the poison ratio on backdoor mitigation performance. As shown in Figure~\ref{fig:prBERTBase}, the ASR gradually rises as the poison ratio increases, yet it remains within an acceptable range for all attacks. Meanwhile, the CACC fluctuates within a small range with the increasing poison ratio. These results indicate that \textbf{MuScleLoRA can maintain satisfactory backdoor mitigation performance under varying poison ratios}. 

More results and analysis of backdoor mitigation performance on $\text{BERT}_{\text{Large}}$ and RoBERTa are provided in Appendix~\ref{appenSec:performMitigation}, where MuScleLoRA consistently achieves the optimal ASRs, surpassing other baselines significantly.

\subsection{Ablation Study}\label{sec:ablation}
\begin{table*}
  \resizebox{\linewidth}{!}{ 
  \begin{tabular}{ccccccccccccc}
  \toprule
  \multirow{2}{*}{Dataset} & \multirow{2}{*}{Method} & \multicolumn{3}{c}{Strategies}             & \multicolumn{2}{c}{Badnets} & \multicolumn{2}{c}{Addsent} & \multicolumn{2}{c}{HiddenKiller} & \multicolumn{2}{c}{StyleBkd} \\
                           &                       & MS         & LR           & GA           & CACC$\uparrow$         & ASR$\downarrow$          & CACC$\uparrow$         & ASR$\downarrow$          & CACC$\uparrow$            & ASR$\downarrow$            & CACC$\uparrow$          & ASR$\downarrow$          \\ \midrule
  \multirow{6}{*}{SST-2}   & Vanilla               & $\times$      & $\times$      & $\times$      & 91.27        & 94.63        & 90.99        & 99.89        & 91.10            & 93.53          & 91.71          & 77.19        \\
                           & \textbf{MuscleLoRA}            & $\checkmark$ & $\checkmark$ & $\checkmark$ & 86.54        & \textbf{12.94}        & 86.77        & \textbf{18.97}        & 87.64           & \textbf{25.11}          & 87.81         & \textbf{33.22}        \\
                           & w/o MS, GA                  & $\times$ &  $\checkmark$     & $\times$      & 89.07        & 65.57        & 88.96        & 96.16        & 88.58           & 52.96          & 88.91         & 57.24        \\
                           & w/o MS, LR               & $\times$      & $\times$      & $\checkmark$ & 91.37        & 90.13        & 90.06        & 100          & 90.39           & 86.40           & 91.21         & 70.61        \\
                           & w/o GA     & $\checkmark$ & $\checkmark$ & $\times$      & 87.64        & 42.76        & 87.75        & 75.22        & 86.88           & \underline{37.39}          & 87.26         & 54.17        \\
                           & w/o MS     & $\times$ &  $\checkmark$     & $\checkmark$ & 83.20         & \underline{24.89}        & 82.81        & \underline{20.06}        & 81.77           & 38.92          & 80.62         & \underline{45.83}        \\
                            \midrule
  \multirow{6}{*}{AG}      & Vanilla               & $\times$      & $\times$      & $\times$      & 92.80         & 51.25        & 92.75        & 100          & 92.78           & 99.47          & 92.06         & 87.59        \\
                           & \textbf{MuscleLoRA}            & $\checkmark$ & $\checkmark$ & $\checkmark$ & 87.74        & \textbf{2.35}         & 87.72         & \textbf{3.90}         & 86.05           & \underline{17.02}          & 87.97         & \textbf{2.67}         \\
                           & w/o MS, GA                  &  $\times$ &$\checkmark$       & $\times$      & 89.59        & 3.28         & 89.05        & 100          & 89.01           & 98.16          & 88.39         & 77.76        \\
                           & w/o MS, LR               & $\times$      & $\times$      & $\checkmark$ & 92.24        & 63.13        & 92.65        & 100          & 93.10            & 99.98          & 93.01         & 92.19        \\
                           & w/o GA     & $\checkmark$ & $\checkmark$ & $\times$      & 89.92        & 2.63         & 89.55        & 99.94        & 89.55           & 97.54          & 89.13         & 71.78        \\
                           & w/o MS     & $\times$  & $\checkmark$      & $\checkmark$ & 84.32        & \underline{2.39}         & 84.85        & \underline{4.07}         & 84.26           & \textbf{8.18}           & 86.38         & \underline{2.79}         \\ \bottomrule
  \end{tabular}
  }
  \caption{The results of ablation experiments when adopting $\text{BERT}_{\text{Base}}$ as the target LM on SST-2 and Agnews. Bold values indicate optimal ASRs and underlined values indicate suboptimal ASRs.}
  \label{tab:ablationBERTBase}
\end{table*}

\begin{figure*}
  \centering
  \begin{subfigure}{0.24\linewidth}
      \centering
      \includegraphics[width=\linewidth]{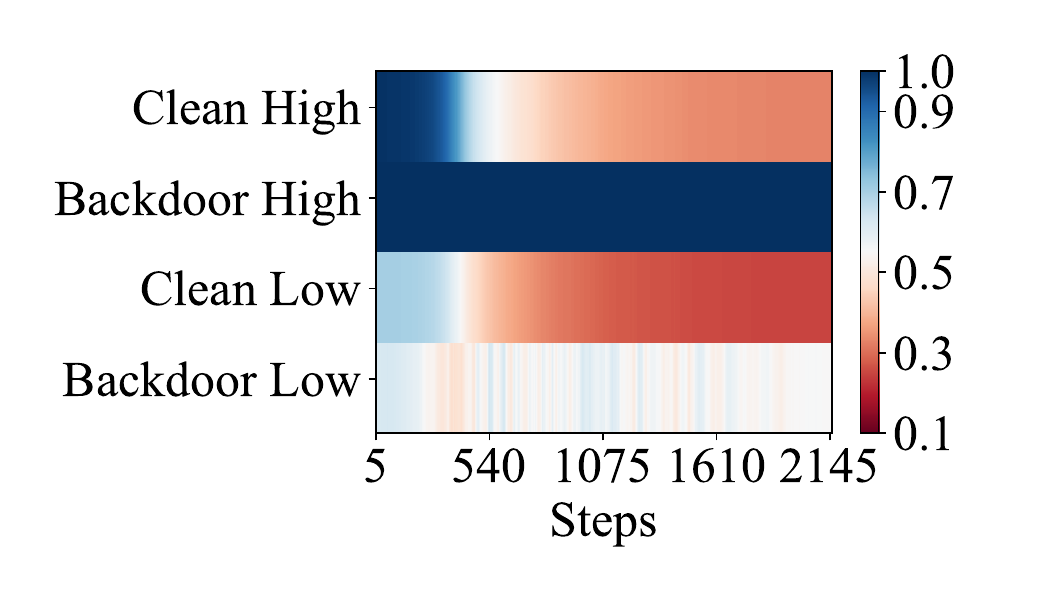}
      \caption{w/o MS, GA}
      \label{subfig:BERTBaseLora}
  \end{subfigure}
  % \hfill
  \begin{subfigure}{0.24\linewidth}
      \centering
      \includegraphics[width=\linewidth]{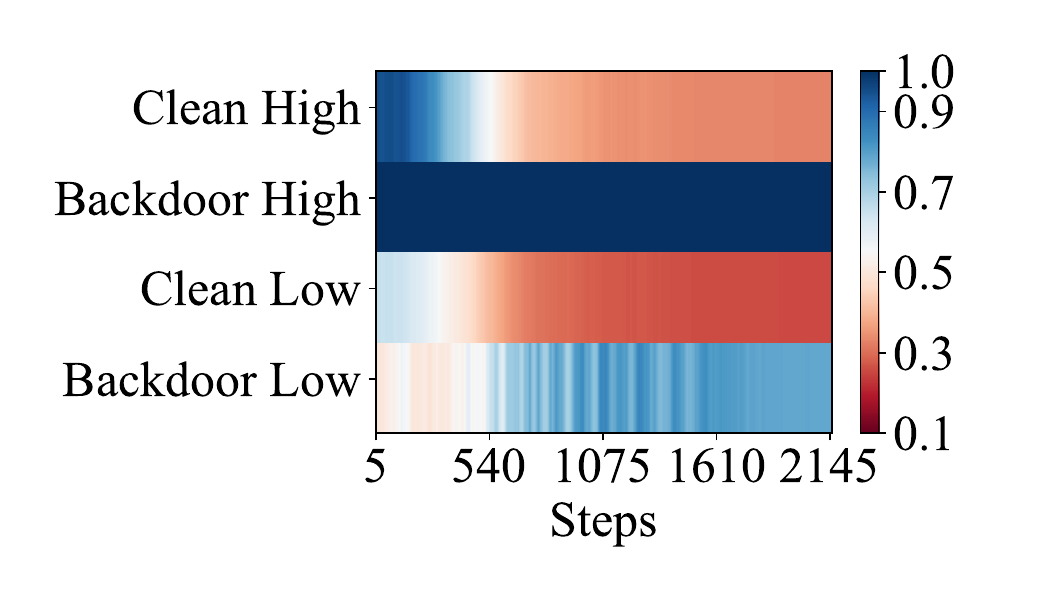}
      \caption{w/o GA}
      \label{subfig:BERTBaseMSLR}
  \end{subfigure}
  \begin{subfigure}{0.24\linewidth}
    \centering
    \includegraphics[width=\linewidth]{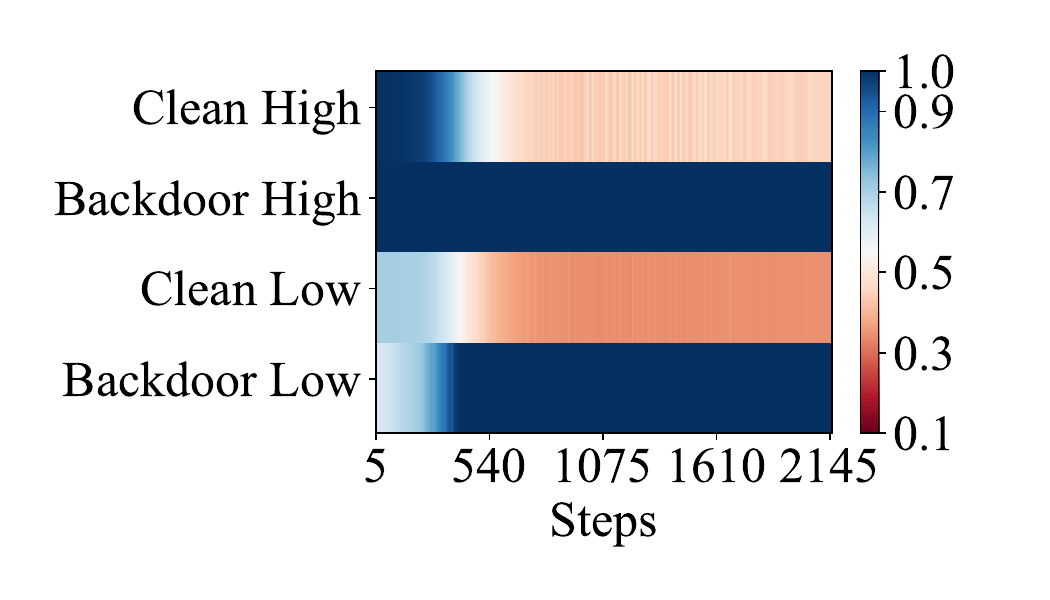}
    \caption{w/o MS }
    \label{subfig:BERTBaseGALoRA}
\end{subfigure}
\begin{subfigure}{0.24\linewidth}
  \centering
  \includegraphics[width=\linewidth]{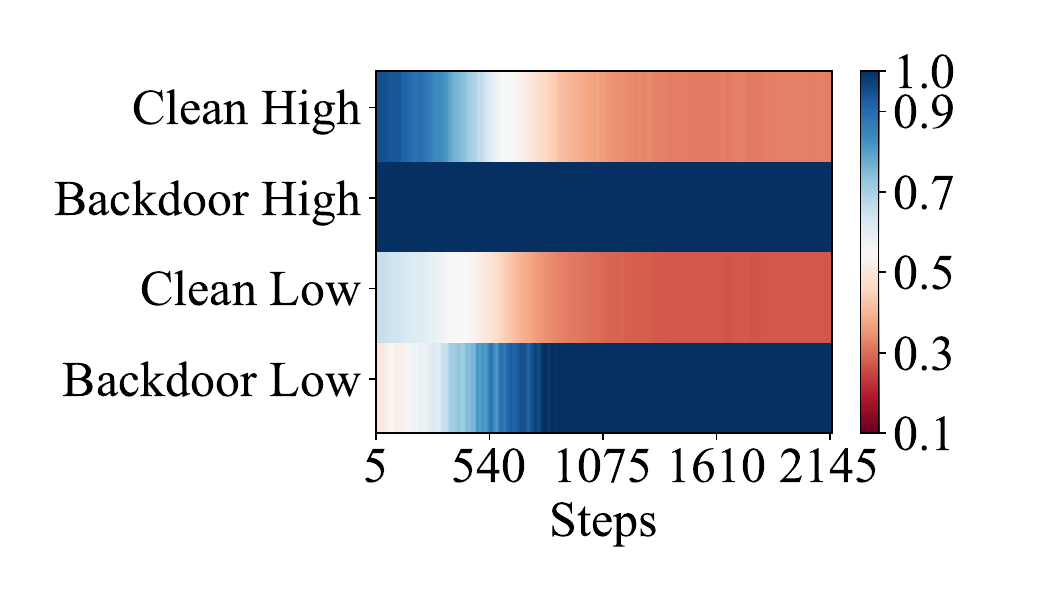}
  \caption{MuScleLoRA}
  \label{subfig:BERTBaseGALoRAMSLR}
\end{subfigure}
  \caption{Relative errors of MuScleLoRA and its ablation methods when adopting $\text{BERT}_{\text{Base}}$ as the target LM on Badnets poisoned SST-2 during training. \vspace{-0.4cm}}
  \label{fig:fourierAnalysisBERTBase}
\end{figure*}

We examine the contributions of three strategies in MuScleLoRA to the results, i.e., multiple radial scalings (MS), low-rank adaptation (LR), and gradient alignment (GA). The ablation results on $\text{BERT}_{\text{Base}}$ shown in Table~\ref{tab:ablationBERTBase} indicate that when only deploying low-rank adaptation, i.e., the LoRA baseline, the ASR drops nearly 20\% on SST-2 but nearly remains unchanged on Agnews. Similarly, utilizing solely gradient alignment yields nearly minimal changes in ASR across both datasets. This suggests that aligning the gradient to clean auxiliary data, without additional defense strategies, fails to mitigate the impact of the poisoned gradient.

Compared to employing a single strategy, integrating multiple radial scalings with low-rank adaptation results in a lower ASR than the LoRA baseline, potentially achieving suboptimal ASR. Additionally, utilizing gradient alignment to low-rank adaptation can reduce ASR for several attacks to suboptimal levels, while achieving the optimal ASR on AGnews. Yet, \textbf{without multiple radial scalings to enhance learning by downscaling clean mapping, CACC drops to an unacceptable level} in this scenario. Consequently, MuScleLoRA combines three strategies, generally achieving the lowest ASR while maintaining acceptable CACC. 

More ablation results on other LMs are provided in Appendix~\ref{appenSec:ablation}, demonstrating that combining three strategies can achieve optimal performance.

\subsection{Fourier Analyses}\label{sec:fourierAnalysis}
To explain the mechanisms of MuScleLoRA in the frequency space, Fourier analyses are conducted on MuScleLoRA and its ablation methods. The results on $\text{BERT}_{\text{Base}}$ are shown in Figure~\ref{fig:fourierAnalysisBERTBase}. More results on other LMs are provided in Appendix~\ref{appenSec:fourierAnalysis}.

Compared to no defense deployment shown in Figure~\ref{subfig:reBadnetsVanilla}, MuScleLoRA and its ablation methods impede the convergence of low-frequency-dominated backdoor mapping. However, as shown in Figure~\ref{subfig:BERTBaseMSLR}, despite multiple radial scalings expediting the convergence of clean mapping and further hindering the learning process of backdoor mapping compared to LoRA baseline, backdoor mapping still exhibits partial convergence. These phenomena indicate that  \textbf{straightforward model capacity reduction with PET methods fails to effectively defend against complex triggers}. Conversely, as shown in Figure~\ref{subfig:BERTBaseGALoRA}, when aligning gradients to clean auxiliary data in the absence of radial scalings, the convergence of backdoor mapping is effectively hindered, but at the expense of underfitting clean mapping. Therefore, as shown in Figure~\ref{subfig:BERTBaseGALoRAMSLR}, MuScleLoRA integrates multiple scalings to enhance the learning of clean mapping, facilitating the balance between backdoor mitigation and satisfactory performance in downstream tasks. 

\subsection{Performance on LLMs}\label{sec:llama2Result}

\begin{table*}[t]
  \centering
  \resizebox{\linewidth}{!}{ 
  \begin{tabular}{cccccccccc}
    \toprule
    \multirow{2}{*}{Model}     & \multirow{2}{*}{Defense} & \multicolumn{2}{c}{Badnets}      & \multicolumn{2}{c}{Addsent}      & \multicolumn{2}{c}{HiddenKiller} & \multicolumn{2}{c}{StyleBkd}     \\
                               &                          & CACC$\uparrow$ & ASR$\downarrow$ & CACC$\uparrow$ & ASR$\downarrow$ & CACC$\uparrow$ & ASR$\downarrow$ & CACC$\uparrow$ & ASR$\downarrow$ \\ \midrule
    \multirow{6}{*}{GPT2-XL}   & Vanilla                  & 94.45          & 81.91           & 94.45          & 100             & 93.57          & 93.42           & 94.23          & 99.67           \\
                               & LoRA                     & 90.28          & 66.56           & 86.33          & 96.82           & 85.01          & 66.02           & 89.62          & 88.82           \\
                               & Prefix                   & 84.24          & 83.33           & 83.36          & 74.78           & 82.10           & 65.13           & 84.90          & 96.16           \\
                               & ONION                    & 88.80          & 42.21           & 89.07          & 85.42           & 83.53          & 92.00           & 85.67          & 99.56           \\
                               & STRIP                    & 93.90           & 49.12           & 92.81          & 96.71           & 91.54          & 92.10           & 92.59          & 97.37           \\
                               & \textbf{MuscleLoRA}               & 90.94          & \textbf{17.98}           & 91.10          & \textbf{9.32}            & 90.17          & \textbf{18.42}           & 90.94          & \textbf{34.10}           \\ \midrule
    \multirow{6}{*}{$\text{Llama2}_{\text{7B}}$} & Vanilla                  & 95.39          & 98.13           & 94.72          & 100             & 96.05          & 96.05           & 96.43          & 98.58           \\
                               & LoRA                     & 96.21          & 16.56           & 59.39          & 93.53           & 94.45          & 78.07           & 95.61          & 93.86           \\
                               & Prefix                   & 93.79          & 48.58           & 92.52          & 56.91           & 92.42          & 60.20            & 93.52          & 96.05           \\
                               & ONION                    & 90.66          & 40.78           & 91.65          & 85.74           & 86.27          & 96.05           & 88.91          & 97.80            \\
                               & STRIP                    & 95.99          & 57.24           & 95.66          & 97.48           & 91.71          & 94.29           & 95.44          & 95.18           \\
                               & \textbf{MuscleLoRA}               & 94.62          & \textbf{10.64}           & 94.07          & \textbf{13.92}           & 94.62          & \textbf{26.86}           & 94.73          & \textbf{39.03}           \\ \bottomrule
  \end{tabular}
  }
  \caption{Backdoor mitigation performance of MuScleLoRA, PET baselines, and post-training end-to-end baselines when adopting GPT2-XL and $\text{Llama2}_{\text{7B}}$ on SST-2. Bold values indicate optimal ASRs.}
  \label{tab:defensePerformanceLLM}
\end{table*}

Since PET emerges as a novel fine-tuning paradigm for LLMs, we extend MuScleLoRA to GPT2-XL and $\text{Llama2}_{\text{7B}}$ for classification tasks, which focuses specifically on the vertical sentiment analysis task on SST-2. Training-stage end-to-end baselines, such as CUBE, BKI, and RAP, require high-computational retraining, rendering such strategies impractical for the backdoor defense of LLMs. Therefore, we only compare the backdoor mitigation performance of MuScleLoRA, PET baselines, and post-training end-to-end baselines on GPT2-XL and $\text{Llama2}_{\text{7B}}$, as presented in Table~\ref{tab:defensePerformanceLLM}.

As model capacity increases, StyleBkd achieves comparable ASRs to other attack methods, bridging the performance gap presented in Table~\ref{tab:ablationBERTBase}. Besides, due to the significant capacity increase of LLMs, PET baselines and end-to-end baselines struggle to effectively counter four attack methods. Notably, \textbf{MuScleLoRA consistently achieves the lowest ASR for all attacks}. Although MuScleLoRA reduces the ASR against StyleBkd to 34.10\% for GPT2-XL and to 39.03\% for $\text{Llama2}_{\text{7B}}$, which may be attributed to the higher frequency leading by stealthy nature of StyleBkd mentioned in Section~\ref{sec:mainResult}, it still significantly outperforms other baselines. Additionally, given the extensive model capacity of $\text{Llama2}_{\text{7B}}$, the decrease in CACC attributed to low-rank adaptation and gradient alignment can be deemed negligible. These findings indicate the potential for deploying MuScleLoRA in instruction-based fine-tuning of LLMs~\citep{zhang2023instruction}.

\subsection{More Comprehensive Analysis}
We are interested in analyzing the impact of different hyperparameters of MuScleLoRA and the influence of out-of-distribution (OOD) data as the clean auxiliary data. Experimental results indicate that MuScleLoRA is relatively insensitive to some hyperparameters. Additionally, OOD data have a subtle impact on the performance of MuScleLoRA. Due to the page limit, we present the detailed analysis in Appendix~\ref{appenSec:hyparasen} and Appendix~\ref{appenSec:ood}.

\section{Conclusions}
In this paper, we conduct Fourier analyses to investigate the mechanisms of backdoor learning, revealing a notable inclination towards lower frequencies in backdoor mapping compared to clean mapping. Inspired by this observation, we proposed MuScleLoRA, a general backdoor defense method. By downscaling in the frequency space, MuScleLoRA encourages LMs to prioritize the learning of relatively high-frequency clean mapping, consequently mitigating the learning backdoor mapping. Experimental results show the efficacy of MuScleLoRA in defending against diverse backdoor attacks. Notably, MuScleLoRA exhibits generality across various backbone LMs, including BERT, RoBERTa, GPT2-XL, and Llama2.

\section*{Limitations}
Our approach has limitations in two main aspects. First, our method only focuses on the scenario where the defender trains the target LM on the attacker-released poisoned dataset. Other scenarios, such as fine-tuning the poisoned LM on the clean dataset or, more rigorously, fine-tuning the poisoned LM on the poisoned dataset, need further exploration. Second, the scaling factor vector $S$ is relative to the model structure and capacity, requiring pre-training to determine the suitable $S$.

\section*{Ethics Statement}
We propose a general backdoor defense method named MuScleLoRA, designed for scenarios where the defender trains the target LM on the attacker-released poisoned dataset. As all experiments are conducted on publicly available datasets and publicly available models, we believe that our proposed defense method poses no potential ethical risk. 

Our created artifacts are intended to provide researchers or users with a tool for acquiring clean language models from backdoor poisoned datasets. All use of existing artifacts is consistent with their intended use in this paper.

\section*{Acknowledgements}
This work is partially supported by the National Key R\&D Program of China under No. 2023YF3303800 and the Joint Funds of the National Natural Science Foundation of China under No.U21B2020.

\bibliography{arxiv}

\appendix

\section{Filtering-based Fourier Transformation}
\label{appenSec:fbft}
In this section, we provide the detailed processes of filtering-based Fourier transformation used to extract the low-frequency and high-frequency parts of the target-LM-fitted and ground-truth mappings. 

We denote the training dataset of the target LM as $\{x_i, t_i\}_{i=1}^{N} = (X, T)$, where $x_i = \{x_i^1, \cdots, x_i^{L}\}$, $L$, $t_i \in \mathbb{R}^{C}$, $X = \{x_1; \cdots; x_N\} \in \mathbb{R}^{N \times L}$, and $T = \{t_1; \cdots; t_N\} \in \mathbb{R}^{N \times C}$ denote the input ids of the input text with length $L$, the ground-truth one-hot label, the input matrix, and the label matrix, respectively. Notably, LMs often convert discrete input ids into continuous embeddings, i.e., $e = \mathcal{E}(x), e \in \mathbb{R}^{L \times d}$, where $\mathcal{E}$ denotes the embedding layer of the target LM and $d$ denotes the embedding dimension. Besides, embedding updates during training typically exhibit small magnitudes. For simplicity, \textbf{we assume that the embedding of each input id remains unchanged throughout training}. Consequently, the mapping fitted by the target LM can be illustrated as Equation~\ref{equ:lmMapping}, where $y \in \mathbb{R}^C$, $Y = \{y_1; \cdots; y_N\} \in \mathbb{R}^{N \times C}$, and $E = \{e_1; \cdots; e_N\} \in \mathbb{R}^{N \times L \times d}$ denote the output logits, the matrix of output logits, and the tensor of input embeddings, respectively.
\begin{equation}
  \label{equ:lmMapping}
  \begin{aligned}
      \mathcal{F} : \mathbb{R}^{L \times d} &\to \mathbb{R}^{C}, \\
      \mathcal{F}  (e) &= y, \\
      \mathcal{F} (E) & = Y.
  \end{aligned}
\end{equation}

Similarly, the ground-truth mapping utilizing the same embedding layer is illustrated as Equation~\ref{equ:gtMapping}.
\begin{equation}
  \label{equ:gtMapping}
  \begin{aligned}
      \mathcal{T} &: \mathbb{R}^{L \times d} \to \mathbb{R}^{C}, \\
      \mathcal{T} & (e) = t, \\
      \mathcal{T} & (E) = T.
  \end{aligned}
\end{equation}

Practically, when $C > 1$, $\mathcal{F}$ represents the high-dimensional mapping. In such scenarios, employing the high-dimensional discrete Fourier transformation incurs significant computational overhead, posing challenges for real dataset analysis. Therefore, we opt for a pragmatic approach by partitioning the frequency space into two segments, i.e., the low-frequency part with $| \xi | \le \xi_0$ and the high-frequency part with $| \xi | > \xi_0$, to decompose the mapping into the low-frequency part and high-frequency part, respectively. Specifically, we denote the Fourier transformation of $\mathcal{F}$ as $\hat{\mathcal{F}}$ and then decompose $\hat{\mathcal{F}}$ by the indicator $\mathcal{I}(| \xi | \le \xi_0)$ that indicate the low-frequency part in the frequency space, which is illustrated as Equation~\ref{equ:fourierMappingDecomp}.
\begin{equation}
  \label{equ:fourierMappingDecomp}
  \begin{aligned}
    \hat{\mathcal{F}}^{\text{low}} & = \hat{\mathcal{F}} \cdot \mathcal{I}(| \xi | \le \xi_0), \\
    \hat{\mathcal{F}}^{\text{high}} & = \hat{\mathcal{F}} - \hat{\mathcal{F}}^{\text{low}}.
  \end{aligned}
\end{equation}

To further alleviate the computational cost of the high-dimensional indicator, we alternatively apply Gaussian filter $\hat{G}^{\frac{1}{\delta}}(\xi)$ to approximate the indicator $\mathcal{I}(| \xi | \le \xi_0)$, i.e., $\hat{\mathcal{F}}^{\text{low}}  \approx \hat{\mathcal{F}} \cdot \hat{G}^{\frac{1}{\delta}}$, where $\frac{1}{\delta}$ denotes the variance of the Gaussian filter in the frequency space. Consequently, in the corresponding physical space, the low-frequency part $y^{\text{low}, \delta}_{i}$ and high-frequency part $y^{\text{high}, \delta}_{i}$ of the output logits $y_i$ for the entire dataset are obtained through Gaussian convolution, as illustrated in Equation~\ref{equ:gaussianConvLogit}. Here, $G^{\delta}(e_i^{\prime}-e_j^{\prime}) = \mathrm{e}^{\frac{-\| e_i^{\prime}-e_j^{\prime} \|^2}{2 \delta}}$ denotes the corresponding Gaussian filter in the physical space with variance $\delta$, $e_i^{\prime} \in \mathbb{R}^{Ld}$ denotes the flattened vector of the embedding $e_i$, $C_i = \sum_{j=1}^{N} G^{\delta}(e_i^{\prime}-e_j^{\prime})$ denotes the normalization factor, and $G \in \mathbb{R}^{N \times N}, G_{ij} = G^{\delta}(e_i^{\prime}-e_j^{\prime})$ denotes the matrix of Gaussian filters, respectively. Practically, we set $\delta$ to 4.0 to obtain frequency components. \looseness=-1
\begin{equation}
  \label{equ:gaussianConvLogit}
  \begin{aligned}
    y^{\text{low}, \delta}_{i} &= \frac{1}{C_i} \sum_{j=1}^{N}y_j G^{\delta}(e_i^{\prime}-e_j^{\prime}) \\
    & = \frac{1}{C_i} (GY)_i, \\
    y^{\text{high}, \delta}_{i} &= y_i - y^{\text{low}, \delta}_{i} \\
    &= \left(Y - \frac{1}{C_i} (GY) \right)_i. \\
  \end{aligned}
\end{equation}

Same as the analysis of output logits, for ground-truth labels, we can derive their respective frequency components, i.e., $t^{\text{low}, \delta}_{i}$ and $t^{\text{high}, \delta}_{i}$, by Gaussian convolution, as illustrated in Equation~\ref{equ:gaussianConvGT}.
\begin{equation}
  \label{equ:gaussianConvGT}
  \begin{aligned}
    t^{\text{low}, \delta}_{i} &= \frac{1}{C_i} \sum_{j=1}^{N}t_j G^{\delta}(e_i^{\prime}-e_j^{\prime}) \\
    & = \frac{1}{C_i} (GT)_i, \\
    t^{\text{high}, \delta}_{i} &= t_i - t^{\text{low}, \delta}_{i} \\
    &= \left(T - \frac{1}{C_i} (GT) \right)_i. \\
  \end{aligned}
\end{equation}

\section{Detailed Experiment Setup}
In this section, we provide additional setup information for experiments. In Section~\ref{appenSec:datasets}, we provide the detailed statistics of datasets. Subsequently, in Section~\ref{appenSec:defenseBaselines}, we provide comprehensive descriptions of defense baselines. Additionally, we outline detailed trigger settings in Section~\ref{appenSec:triggerSettings}. Besides, Section~\ref{appenSec:hyperparameters} elaborates on hyperparameter settings. Furthermore, Section~\ref{appenSec:usageArtifacts} provides the usage of existing artifacts.

\subsection{Datasets}\label{appenSec:datasets}
The statistics of datasets are presented in Table~\ref{tab:datasets}. Considering the excessive number of samples in Agnews, which could potentially prolong the training process, \textbf{we decided to randomly extract 5,000 samples from each class in the original training dataset}. Consequently, a new training dataset comprising 20,000 samples is synthesized.
\begin{table}
  \centering
  \setlength{\tabcolsep}{2pt}
  \resizebox{\linewidth}{!}{ 
  \begin{tabular}{cccccc}
  \toprule
  \multirow{2}{*}{Dataset} & \multirow{2}{*}{Categories} & \multicolumn{3}{c}{Number of Samples} & \multirow{2}{*}{\makecell{Average \\Length}} \\
                &           & Train      & Test     & Validation    &                                 \\ \midrule
  SST-2        & 2            & 6,920      & 1,821    & 872           & 19.2                            \\
  HSOL         & 2            & 5,823      & 2,485    & 2,485         & 13.2                            \\
  Lingspam     & 2            & 2,604      & 582      & 289           & 695.3                           \\
  Agnews       & 4            & 108,000    & 7,600    & 12,000        & 38.0                              \\ \bottomrule
  \end{tabular}
  }
  \caption{Detailed statistics of datasets.}
  \label{tab:datasets}
\end{table}

\subsection{Defense Baselines}\label{appenSec:defenseBaselines}
% \noindent \textbf{PET baselines.} 
\paragraph{\textbf{PET baselines.} } 
PET baselines reduce model capacity by freezing the original weights of the LM and inserting tunable PET modules with a small number of parameters, constraining the model to focus on clean tasks \citep{zhu2022moderate}. LoRA~\citep{hu2021lora} inserts parallel low-rank decompositions as the tunable module. Adapter~\citep{houlsby2019parameter} inserts a sequential linear layer as the tunable module. Prefix-Tuning~\citep{li2021prefix} inserts a sequence of continuous task-specific vectors as the tunable module. \looseness=-1

\paragraph{\textbf{ONION.}}  Based on the observation that inserting trigger words into original text results in a notable increase in perplexity, ONION~\citep{qi2021onion} utilizes GPT-2 to quantify the contribution of each word in the original text to the perplexity and detect high-contributing words as the trigger words.

\paragraph{\textbf{STRIP.}}  Based on the observation that clean text is more sensitive to perturbations than poisoned text, STRIP~\citep{gao2021design} employs random word replacement to perturb input text, subsequently identifying poisoned text by analyzing discrepancy in the entropy of output logits.

\paragraph{\textbf{RAP.}}  Similar to STRIP, RAP~\citep{yang2021rap}  discerns poisoned input texts based on their sensitivity to perturbations. RAP reconfigures the embedding layer to incorporate a robust-aware perturbation to be introduced into input texts, which significantly alters the logits of clean texts while minimally affecting poisoned samples. 

\paragraph{\textbf{BKI.}}  Similar to ONION, BKI~\citep{chen2021mitigating} quantifies the contribution of each word in the original text of the training dataset to the output logits to detect high-contributing words as the trigger words.

\paragraph{\textbf{CUBE.}}  Based on the observation that poisoned samples frequently manifest as outliers in the feature space, CUBE~\citep{cui2022unified} clusters samples in the training dataset to identify outliers as the poisoned samples.

\subsection{Trigger Settings}\label{appenSec:triggerSettings}
For Badnets, following the settings of~\citet{kurita2020weight}, we insert 4 rare words, i.e., \textit{cf}, \textit{mn}, \textit{bb}, and \textit{tq}, into random positions within the original text. For Addsent, following the settings of~\citet{dai2019backdoor}, we insert a predefined sentence, i.e., \textit{I watch this 3D movie}, into a random position within the original text. For HiddenKiller, following the settings of~\citet{qi2021hidden}, we adopt \textit{( ROOT ( S ( SBAR ) ( , ) ( NP ) ( VP ) ( . ) ) ) EOP} as the trigger syntax. We then paraphrase the entire original text into trigger syntax for the sentence-level datasets: SST-2, HSOL, and Agnews. Additionally, for the paragraph-level dataset Lingspam, each sentence in the original text is paraphrased into trigger syntax. For StyleBkd, following the settings of~\citet{qi2021mind}, we choose \textit{bible} text style as the trigger style. Similar to HiddenKiller, we paraphrase the entire original text into trigger style for the sentence-level datasets: SST-2, HSOL, and Agnews, while every sentence in the original text is paraphrased into trigger style for the paragraph-level dataset Lingspam.

\subsection{Hyperparameters}\label{appenSec:hyperparameters}
Notably, compared to the meticulous hyperparameter design by \citet{zhu2022moderate} tailored for different attacks, \textbf{we unify hyperparameters against diverse attacks for each specific LM}. Specifically, following the observation that reducing the training epochs can mitigate backdoor learning~\citep{zhu2022moderate}, we set training epochs to 10 for BERT and RoBERTa, and 5 for LLMs. Similarly, we set learning rate to $2 \times 10^{-5}$ for BERT and RoBERTa, and $10^{-5}$ for LLMs. Additionally, considering the extensive model capability of LLMs, the number of clean auxiliary data for LLMs is set to 128 whereas it is set to 96 for BERT and RoBERTa. Furthermore, $\mu_{\max}$ is configured to 0 for LLMs and 0.1 for BERT and RoBERTa. The batch size is defined as 16 for LLMs and 32 for BERT and RoBERTa. For PET baselines, the bottleneck dimensions are uniformly set to 8 for BERT and RoBERTa and 2 for LLMs. Finally, detailed settings of the scaling factor vector $S$ are presented in Table~\ref{tab:scalingfactors}, and the bandwidth $r$ of each $A_i$ in radial scalings is specified as only \textbf{1}. All experiments are conducted on NVIDIA GeForce RTX 3090 with 24GB memory.

\begin{table}
  \centering
  \small
%   \setlength{\tabcolsep}{2pt}
  % \resizebox{\linewidth}{!}{ 
  \begin{tabular}{cc}
  \toprule
  Model & $S$ \\ \midrule
  $\text{BERT}_{\text{Base}}$ & $[1, 4, 8, 12, 16, 20, 24, 28]$ \\
  $\text{BERT}_{\text{Large}}$ & $[1, 2, 3, 4, 5, 6, 7, 8, 9]$ \\
  $\text{RoBERTa}_{\text{Base}}$ & $[1, 2, 4, 6, 8, 10, 12, 14, 16]$ \\
  $\text{RoBERTa}_{\text{Large}}$ & $[1, 2, 3, 4, 5, 6, 7, 8, 9]$ \\
  $\text{Llama2}_{\text{7B}}$ & $[1, 2, 3, 4]$ \\
  \bottomrule
  \end{tabular}
  % }
  \caption{Detailed settings of scaling factor vector $S$. }
  % \vspace{-0.5cm}
  \label{tab:scalingfactors}
\end{table}

\subsection{Usage of Existing Artifacts}\label{appenSec:usageArtifacts}
For conducting backdoor attacks and end-to-end defense baselines, we employ OpenBackdoor~\citep{cui2022unified}, an open-source framework for textual backdoor learning. The detailed process of MuScleLoRA is implemented within the framework of PyTorch~\citep{paszke2019pytorch}, an open-source library for deploying deep learning. For implementing PET algorithms, we utilize Huggingface-PEFT~\citep{sourab2022peft}, an open-source library for HuggingFace-transformers-based PET methods of LMs, and Opendelta~\citep{hu2023opendelta}, another open-source library dedicated to PET methods of LMs. For LMs, we adopt BERT, RoBERTa, and $\text{Llama2}_{\text{7B}}$ from Huggingface transformers\footnote{\url{https://github.com/huggingface/transformers}}. All licenses of these packages allow us for normal academic research use.

\section{Additional Experimental Results and Analyses}
In this section, we provide additional experimental results and analyses. Section~\ref{appenSec:performMitigation} provides the backdoor mitigation performance on $\text{BERT}_{\text{Large}}$, $\text{RoBERTa}_{\text{Base}}$, and $\text{RoBERTa}_{\text{Large}}$. Subsequently, we conduct the ablation studies on the three strategies in MuScleLoRA when adopting $\text{BERT}_{\text{Large}}$, $\text{RoBERTa}_{\text{Base}}$, or $\text{RoBERTa}_{\text{Large}}$ as the target LM in Section~\ref{appenSec:ablation}, conduct Fourier analyses on $\text{BERT}_{\text{Large}}$ and $\text{Llama2}_{\text{7B}}$ to explain the mechanisms of MuScleLoRA in Section~\ref{appenSec:fourierAnalysis}, analyze the sensitivity on hyperparameters in Section~\ref{appenSec:hyparasen}, and analyze the impact of out-of-distribution (OOD) data in gradient alignment in Section~\ref{appenSec:ood}.

\begin{table*}
  % \resizebox{\linewidth}{!}{ 
  \centering
  \small
  \setlength{\tabcolsep}{5pt}
  \begin{tabular}{ccccccccccc}
  \toprule
  \multirow{2}{*}{Dataset} & \multirow{2}{*}{Model}                           & \multirow{2}{*}{Defense} & \multicolumn{2}{c}{Badnets} & \multicolumn{2}{c}{Addsent} & \multicolumn{2}{c}{HiddenKiller} & \multicolumn{2}{c}{StyleBkd} \\
                           &                                                  &                          & CACC$\uparrow$         & ASR$\downarrow$          & CACC$\uparrow$         & ASR$\downarrow$          & CACC$\uparrow$            & ASR$\downarrow$            & CACC$\uparrow$          & ASR$\downarrow$          \\ \midrule
  \multirow{15}{*}{SST-2}  & \multirow{5}{*}{$\text{BERT}_{\text{Large}}$}    & Vanilla                  & 92.91        & 93.64        & 92.97        & 100          & 92.64           & 90.24          & 93.30         & 78.51        \\
                           &                                                  & LoRA                     & 91.98        & 31.14        & 91.27        & 84.87        & 91.54           & 42.21          & 90.50         & 66.67        \\
                           &                                                  & Adapter                  & 89.73        & 40.57        & 88.85        & 70.17        & 89.51           & 42.98          & 89.07         & 64.14        \\
                           &                                                  & Prefix            & 92.42        & 37.06        & 92.04        & 99.56        & 92.59           & 67.98          & 91.93         & 57.90        \\
                           &                                                  & \textbf{MuscleLoRA}               & 91.21        & \textbf{14.80}        & 90.71        & \textbf{27.30}        & 90.99           & \textbf{17.54}          & 89.62         & \textbf{21.16}        \\ \cmidrule{2-11} 
                           & \multirow{5}{*}{$\text{RoBERTa}_{\text{Base}}$}  & Vanilla                  & 94.39        & 95.61        & 94.17        & 99.89        & 93.13           & 93.86          & 94.67         & 99.34        \\
                           &                                                  & LoRA                     & 92.09        & 26.54        & 91.87        & 63.05        & 90.99           & 38.60          & 91.54         & 67.33        \\
                           &                                                  & Adapter                  & 91.43        & 57.46        & 88.69        & 62.39        & 91.49           & 33.77          & 90.45         & 69.96        \\
                           &                                                  & Prefix            & 91.98        & 85.19        & 91.98        & 100          & 90.94           & 62.94          & 92.36         & 94.96        \\
                           &                                                  & \textbf{MuscleLoRA}               & 88.08        & \textbf{13.26}        & 88.91        & \textbf{21.16}        & 89.07           & \textbf{20.28}          & 88.41         & \textbf{20.61}        \\ \cmidrule{2-11} 
                           & \multirow{5}{*}{$\text{RoBERTa}_{\text{Large}}$} & Vanilla                  & 94.29        & 100          & 95.44        & 100          & 93.52           & 90.24          & 94.45         & 99.12        \\
                           &                                                  & LoRA                     & 95.55        & 11.73        & 94.95        & 92.21        & 95.94           & 57.24          & 95.39         & 73.03        \\
                           &                                                  & Adapter                  & 70.01        & 99.78        & 58.81        & 35.52        & 58.10           & 52.19          & 62.55         & 96.05        \\
                           &                                                  & Prefix            & 94.89        & 76.54        & 94.56        & 78.73        & 93.96           & 62.50          & 94.62         & 89.14        \\
                           &                                                  & \textbf{MuscleLoRA}              & 93.30        & \textbf{5.81}         & 93.19        & \textbf{14.47}        & 92.59           & \textbf{10.96}          & 92.48         & \textbf{12.39}        \\ \midrule
  \multirow{15}{*}{AG}     & \multirow{5}{*}{$\text{BERT}_{\text{Large}}$}    & Vanilla                  & 93.71        & 63.86        & 93.56        & 100          & 93.53           & 99.32          & 93.18         & 88.21        \\
                           &                                                  & LoRA                     & 90.67        & 1.68         & 90.55        & 99.81        & 90.32           & 97.21          & 90.21         & 82.99        \\
                           &                                                  & Adapter                  & 90.16        & 3.68         & 89.45        & 66.53        & 89.74           & 91.00          & 88.97         & 36.72        \\
                           &                                                  & Prefix            & 92.39        & 54.81        & 92.54        & 100          & 91.75           & 99.10          & 91.76         & 82.99        \\
                           &                                                  & \textbf{MuscleLoRA}               & 89.58        & \textbf{1.67}         & 89.10        & \textbf{1.70}         & 87.33           & \textbf{28.04 }         & 88.97         & \textbf{12.15}        \\ \cmidrule{2-11} 
                           & \multirow{5}{*}{$\text{RoBERTa}_{\text{Base}}$}  & Vanilla                  & 93.29        & 86.19        & 93.68        & 100          & 93.32           & 99.98          & 93.56         & 91.56        \\
                           &                                                  & LoRA                     & 90.54        & 1.86         & 90.22        & 99.96        & 90.53           & 99.93          & 89.93         & 81.28        \\
                           &                                                  & Adapter                  & 90.60        & 3.40         & 89.85        & 99.98        & 90.39           & 99.70          & 88.96         & 78.77        \\
                           &                                                  & Prefix            & 91.05        & 39.51        & 91.12        & 99.95        & 90.87           & 99.96          & 90.33         & 84.63        \\
                           &                                                  & \textbf{MuscleLoRA}               & 86.89        & \textbf{1.42}         & 86.30        & \textbf{1.35}        & 87.01           & \textbf{19.46}          & 86.78         & \textbf{5.70}         \\ \cmidrule{2-11} 
                           & \multirow{5}{*}{$\text{RoBERTa}_{\text{Large}}$} & Vanilla                  & 93.79        & 96.42        & 93.14        & 100          & 93.66           & 100            & 93.59         & 94.40        \\
                           &                                                  & LoRA                     & 92.14        & 2.21         & 92.24        & 99.96        & 91.96           & 99.90          & 91.63         & 88.44        \\
                           &                                                  & Adapter                  & 91.10        & 2.39         & 91.10        & 99.95        & 90.83           & 99.23          & 90.75         & 72.75        \\
                           &                                                  & Prefix            & 92.34        & 18.60        & 92.18        & 99.96        & 92.21           & 99.98          & 91.82         & 91.12        \\
                           &                                                  & \textbf{MuscleLoRA}               & 90.21        & \textbf{1.85}         & 90.10        & \textbf{4.26}         & 89.64          & \textbf{7.34}           & 90.05         & \textbf{2.30}         \\ \bottomrule
  \end{tabular}
  % }
  \caption{Backdoor mitigation performance of MuScleLoRA and PET baselines when adopting $\text{BERT}_{\text{Large}}$, $\text{RoBERTa}_{\text{Base}}$, or $\text{RoBERTa}_{\text{Large}}$ as the target LM on SST-2 and Agnews. Bold values indicate optimal ASRs.}
  \label{tab:petPerformanceAdditionalModels}
\end{table*}

\begin{table*}
  \resizebox{\linewidth}{!}{ 
  \begin{tabular}{ccccccccccccc}
  \toprule
  \multirow{2}{*}{Model}                           & \multirow{2}{*}{Method} & \multicolumn{3}{c}{Strategies}             & \multicolumn{2}{c}{Badnets} & \multicolumn{2}{c}{Addsent} & \multicolumn{2}{c}{HiddenKiller} & \multicolumn{2}{c}{StyleBkd} \\
                                                   &                       & MS         & LR           & GA           & CACC$\uparrow$         & ASR$\downarrow$          & CACC$\uparrow$         & ASR$\downarrow$          & CACC$\uparrow$            & ASR$\downarrow$            & CACC$\uparrow$          & ASR$\downarrow$          \\ \midrule
  \multirow{6}{*}{$\text{BERT}_{\text{Large}}$}    & Vanilla               & $\times$     & $\times$     & $\times$     & 92.91        & 93.64        & 92.97        & 100          & 92.64           & 90.24          & 93.30         & 78.51        \\
                                                   & \textbf{MuscleLoRA}            & $\checkmark$ & $\checkmark$ & $\checkmark$ & 91.21        & \textbf{14.80}        & 90.72        & \textbf{27.30}        & 90.99           & \textbf{17.54}          & 89.62         & \textbf{21.16}        \\ 
                                                   & w/o MS, GA                  & $\times$ & $\checkmark$     & $\times$     & 91.98        & 31.14        & 91.27        & 84.87        & 91.54           & 42.21          & 90.50         & 66.67        \\
                                                   & w/o MS, LR           & $\times$     & $\times$     & $\checkmark$ & 93.68        & 89.91        & 92.53        & 100          & 91.98           & 86.62          & 92.58         & 74.67        \\
                                                   & w/o GA     & $\checkmark$ & $\checkmark$ & $\times$     & 91.54        & \underline{29.71}        & 90.28        & 75.22        & 90.94           & 44.71          & 89.62         & 54.47        \\
                                                   & w/o MS     & $\times$ & $\checkmark$     & $\checkmark$ & 86.54        & 29.82        & 86.16        & \underline{36.84}        & 87.37           & \underline{27.85}          & 85.94         & \underline{28.29}        \\\midrule                                              
  \multirow{6}{*}{$\text{RoBERTa}_{\text{Base}}$}  & Vanilla               & $\times$     & $\times$     & $\times$     & 94.39        & 95.61        & 94.17        & 99.89        & 93.13           & 93.86          & 94.67         & 99.34        \\
                                                   & \textbf{MuscleLoRA}            & $\checkmark$ & $\checkmark$ & $\checkmark$ & 88.08        & \textbf{13.26}        & 88.91        & \textbf{21.16}        & 89.07           & \textbf{20.28}          & 88.41         & \textbf{20.61}        \\
                                                   & w/o MS, GA                  & $\times$  & $\checkmark$    & $\times$     & 92.09        & 26.54        & 91.87         & 63.05        & 90.99           & 38.60          & 91.54         & 67.33        \\
                                                   & w/o  MS, LR               & $\times$     & $\times$     & $\checkmark$ & 92.86        & 94.30        & 93.46        & 100          & 90.06           & 87.61          & 94.12         & 96.71        \\
                                                   & w/o GA     & $\checkmark$ & $\checkmark$ & $\times$     & 93.30        & \underline{24.45}        & 92.53        & 65.57        & 92.31           & 48.68          & 92.81         & 40.46        \\
                                                   & w/o MS     & $\times$ & $\checkmark$     & $\checkmark$ & 80.72        & 25.22        & 80.45        & \underline{22.92}        & 82.87           & \underline{22.81}          & 84.57         & \underline{23.13}        \\\midrule                                                 
  \multirow{6}{*}{$\text{RoBERTa}_{\text{Large}}$} & Vanilla               & $\times$     & $\times$     & $\times$     & 94.29        & 100          & 95.44        & 100          & 93.52           & 90.24          & 94.45         & 99.12        \\
                                                   & \textbf{MuscleLoRA}            & $\checkmark$ & $\checkmark$ & $\checkmark$ & 93.30        & \textbf{5.81}         & 93.19        & \textbf{14.47}        & 92.59           & \textbf{10.96}          & 92.48         & \textbf{12.39}        \\ 
                                                   & w/o MS, GA                  & $\times$ & $\checkmark$     & $\times$     & 95.55        & 11.73        & 94.95        & 92.21        & 95.94           & 57.24          & 95.39         & 73.03        \\
                                                   & w/o  MS, LR              & $\times$     & $\times$     & $\checkmark$ & 94.95        & 67.21        & 95.44        & 100          & 95.28           & 90.24          & 95.39         & 92.21        \\
                                                   & w/o GA     & $\checkmark$ & $\checkmark$ & $\times$     & 94.84        & 13.05        & 94.40        & 70.83        & 95.28           & 44.74          & 95.72         & 71.49        \\
                                                   & w/o MS     & $\times$ & $\checkmark$     & $\checkmark$ & 89.79        & \underline{10.31}        & 90.39        & \underline{18.75}        & 91.05           & \underline{16.45}          & 91.43         & \underline{16.67}        \\\bottomrule
                                                   
  \end{tabular}
  }
  \caption{The results of ablation experiments on SST-2 when adopting $\text{BERT}_{\text{Large}}$, $\text{RoBERTa}_{\text{Base}}$, or $\text{RoBERTa}_{\text{Large}}$ as the respective target LM. Bold values indicate optimal ASRs and underlined values indicate suboptimal ASRs.}
  \label{tab:ablationAdditionalModels}
\end{table*}

\subsection{Performance of Backdoor Mitigation}\label{appenSec:performMitigation}
We further evaluate the backdoor mitigation performance on $\text{BERT}_{\text{Large}}$, $\text{RoBERTa}_{\text{Base}}$, and $\text{RoBERTa}_{\text{Large}}$. As presented in Table~\ref{tab:petPerformanceAdditionalModels}, similar to the results presented in Table~\ref{tab:petPerformanceBERTBase}, although PET baselines manage to reduce the ASR for Badnets to a relatively low level, they still encounter challenges in effectively defending against other complex triggers. Conversely, \textbf{MuScleLoRA consistently reduces the ASR to the lowest level, surpassing the performance of the three PET baselines significantly}. Moreover, in comparison to the about 4-5\% decrease in CACC when implementing MuScleLoRA on $\text{BERT}_{\text{Base}}$, the decrease in CACC for $\text{BERT}_{\text{Large}}$ and $\text{RoBERTa}_{\text{Large}}$ is negligible. This suggests that \textbf{a larger model capacity can alleviate the reduction in CACC while preserving low ASR when deploying MuScleLoRA}.

Also, we evaluate the backdoor mitigation performance of MuScleLoRA with end-to-end defense baselines on $\text{BERT}_{\text{Large}}$. As presented in Table~\ref{tab:e2ePerformanceBERTLarge}, \textbf{MuScleLoRA achieves the optimal ASRs, surpassing all end-to-end baselines}. 

\begin{table}
  \centering
  \setlength{\tabcolsep}{2pt}
  \resizebox{\linewidth}{!}{ 
  \begin{tabular}{ccccccc}
  \toprule
  \multirow{2}{*}{Defense}             & \multicolumn{2}{c}{Addsent} & \multicolumn{2}{c}{HiddenKiller} & \multicolumn{2}{c}{StyleBkd} \\
                                       & CACC$\uparrow$         & ASR$\downarrow$          & CACC$\uparrow$            & ASR$\downarrow$            & CACC$\uparrow$          & ASR$\downarrow$          \\ \midrule
  Vanilla                              & 92.97        & 100          & 92.64           & 90.24          & 93.30         & 78.51        \\
  ONION                                & 88.14        & 93.09        & 86.27           & 96.16          & 87.48         & 79.56        \\
  BKI                                  & 92.20        & 100          & 91.16           & 92.65          & 92.31         & 81.58        \\
  CUBE                                 & 93.24        & 100          & 92.53           & 21.93          & 91.65         & 31.47        \\
  STRIP                                & 72.43        & 60.64        & 92.09           & 91.67          & 89.17         & 75.76        \\
  RAP                                  & 92.04        & 100          & 90.94           & 92.98          & 87.66         & 69.08        \\
  \textbf{MuScleLoRA} & 90.71        & \textbf{27.30}        & 90.99           & \textbf{17.54}          & 89.62         & \textbf{21.16}        \\ \bottomrule
  \end{tabular}
  }
  \caption{Backdoor mitigation performance of MuScleLoRA and end-to-end baselines when adopting $\text{BERT}_{\text{Large}}$ as the target LM on SST-2. Bold values indicate optimal ASRs.}
  \label{tab:e2ePerformanceBERTLarge}
\end{table}

\begin{figure}
  \centering
  \includegraphics[width=\linewidth]{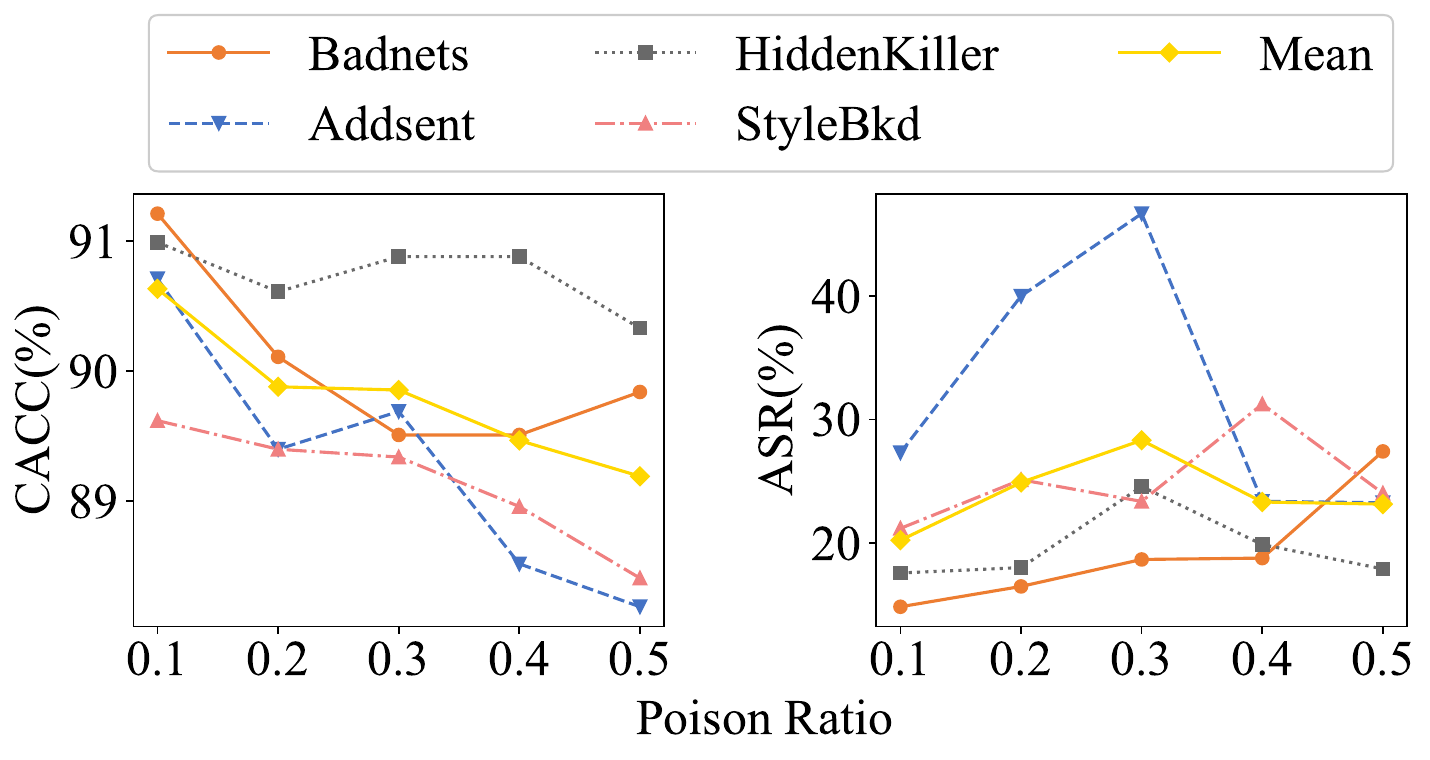}
  \caption{CACC and ASR of MuScleLoRA when adopting $\text{BERT}_{\text{Large}}$ as the target LM on poisoned SST-2 under diverse poison ratios.\vspace{-0.5cm}}
  \label{fig:prBERTLarge}
\end{figure}

Furthermore, experiments are performed to explore the impact of poison ratio on ASR and CACC when adopting $\text{BERT}_{\text{Large}}$ as the target LM. As shown in Figure~\ref{fig:prBERTLarge}, as the poison ratio increases, CACC exhibits a slight decrease, while ASR fluctuates within an acceptable range.

\subsection{Ablation Study}\label{appenSec:ablation}
Additionally, we examine the contribution of three strategies in MuScleLoRA to the performance on SST-2 when adopting $\text{BERT}_{\text{Large}}$, $\text{RoBERTa}_{\text{Base}}$, or $\text{RoBERTa}_{\text{Large}}$ as the target LM, respectively. The results of the ablation analyses are presented in Table~\ref{tab:ablationAdditionalModels}. Similar to the ablation of $\text{BERT}_{\text{Base}}$, solely employing low-rank adaptation or gradient alignment encounters challenges in effectively defending against diverse backdoor attacks. Moreover, the absence of radial scalings leads to a significant drop in CACC. \textbf{Optimal performance is achieved only when all three strategies are combined}.

\subsection{Fourier analyses}\label{appenSec:fourierAnalysis}
\begin{figure*}[h]
  \centering
  \begin{subfigure}{0.24\linewidth}
      \centering
      \includegraphics[width=\linewidth]{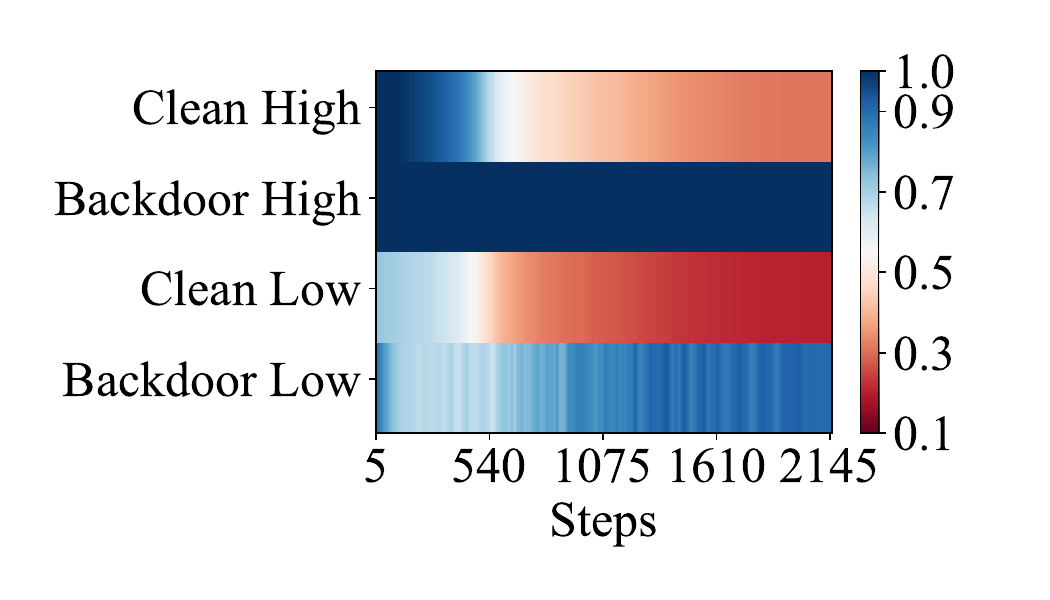}
      \caption{w/o MS, GA}
      \label{subfig:BERTLargeBadnetsLora}
  \end{subfigure}
  \hfill
  \begin{subfigure}{0.24\linewidth}
      \centering
      \includegraphics[width=\linewidth]{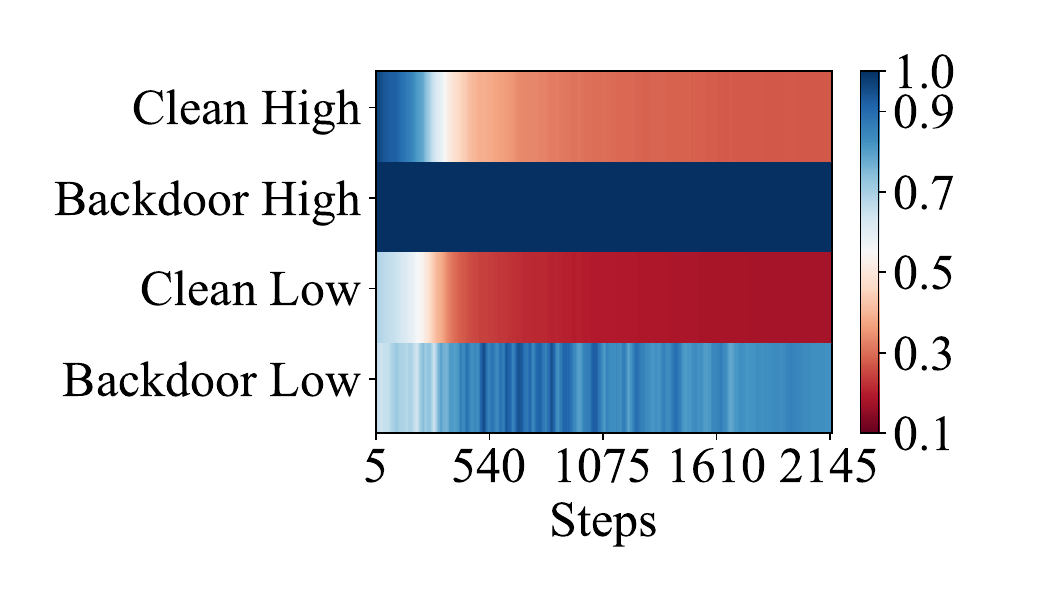}
      \caption{w/o GA}
      \label{subfig:BERTLargeBadnetsMSLR}
  \end{subfigure}
  \begin{subfigure}{0.24\linewidth}
    \centering
    \includegraphics[width=\linewidth]{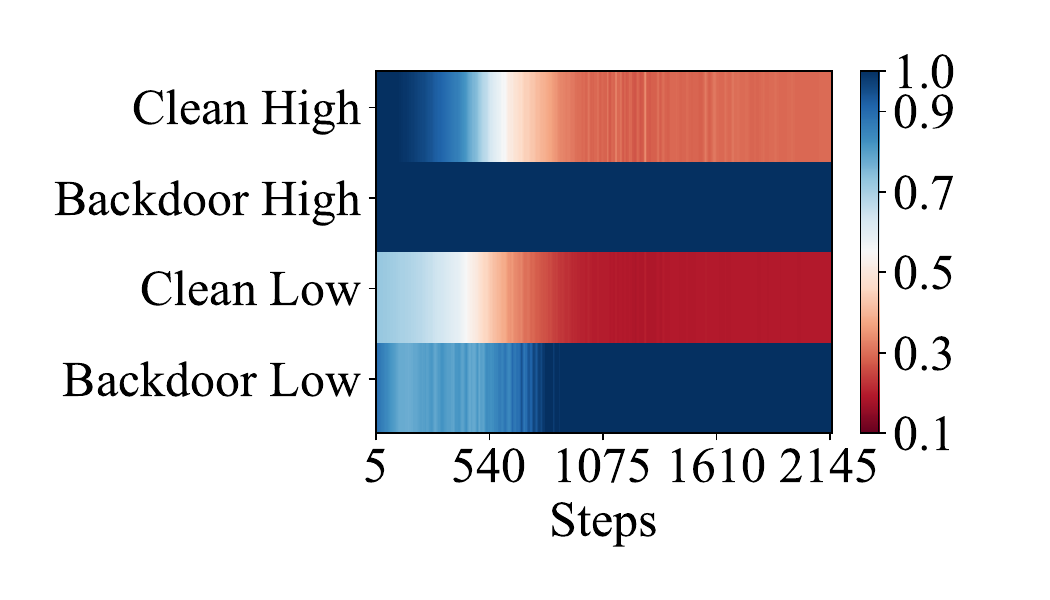}
    \caption{w/o MS }
    \label{subfig:BERTLargeBadnetsGALoRA}
\end{subfigure}
\begin{subfigure}{0.24\linewidth}
  \centering
  \includegraphics[width=\linewidth]{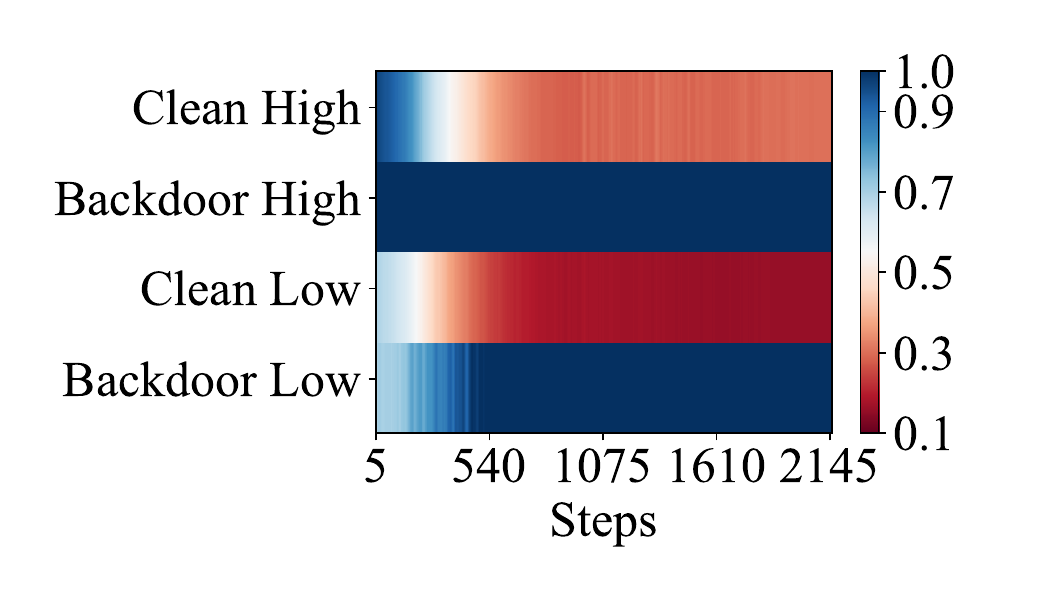}
  \caption{MuScleLoRA}
  \label{subfig:BERTLargeBadnetsGALoRAMSLR}
\end{subfigure}
  \caption{Relative errors of MuScleLoRA and its ablation methods when adopting $\text{BERT}_{\text{Large}}$ as the target LM on Badnets poisoned SST-2 during training.}
  \label{fig:fourierAnalysisBERTLargeBadnets}
\end{figure*}

\begin{figure*}[!t]
  \centering
  \begin{subfigure}{0.24\linewidth}
      \centering
      \includegraphics[width=\linewidth]{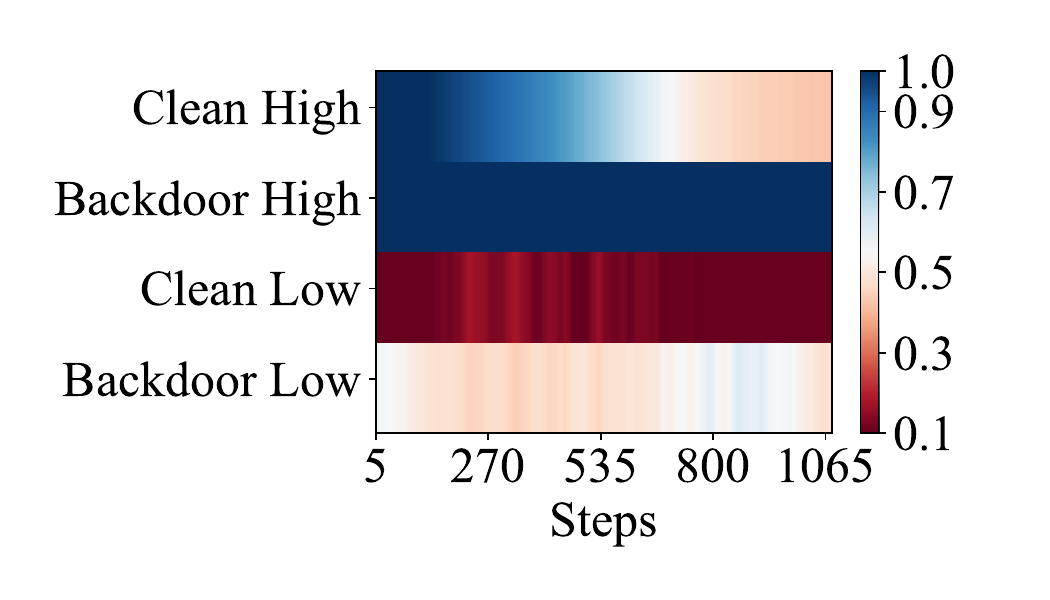}
      \caption{w/o MS, GA}
      \label{subfig:LlamaAddsentsLora}
  \end{subfigure}
  % \hfill
  \begin{subfigure}{0.24\linewidth}
      \centering
      \includegraphics[width=\linewidth]{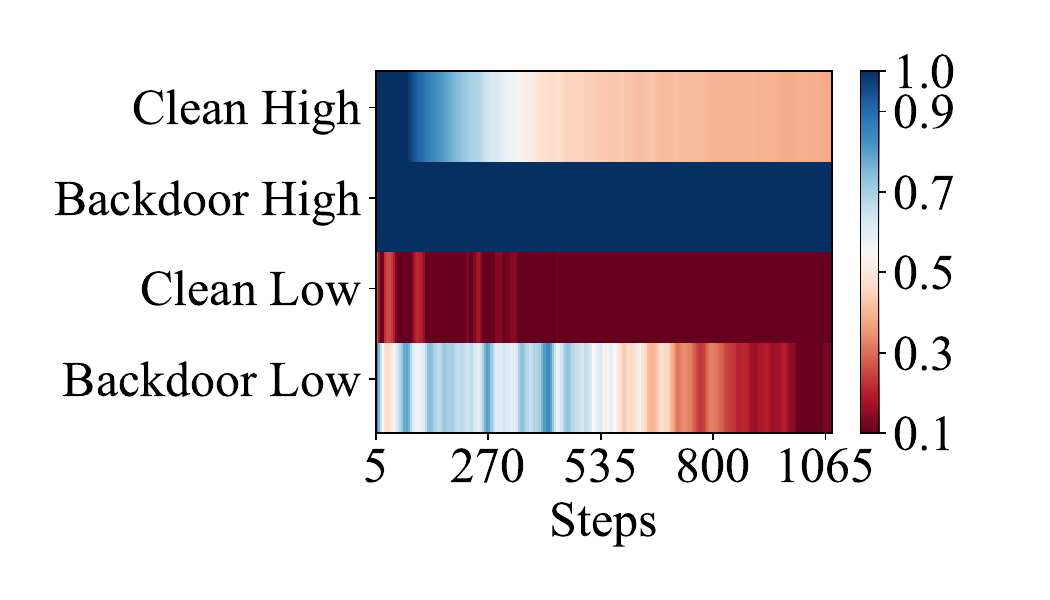}
      \caption{w/o GA}
      \label{subfig:LlamaAddsentsMSLR}
  \end{subfigure}
  \begin{subfigure}{0.24\linewidth}
    \centering
    \includegraphics[width=\linewidth]{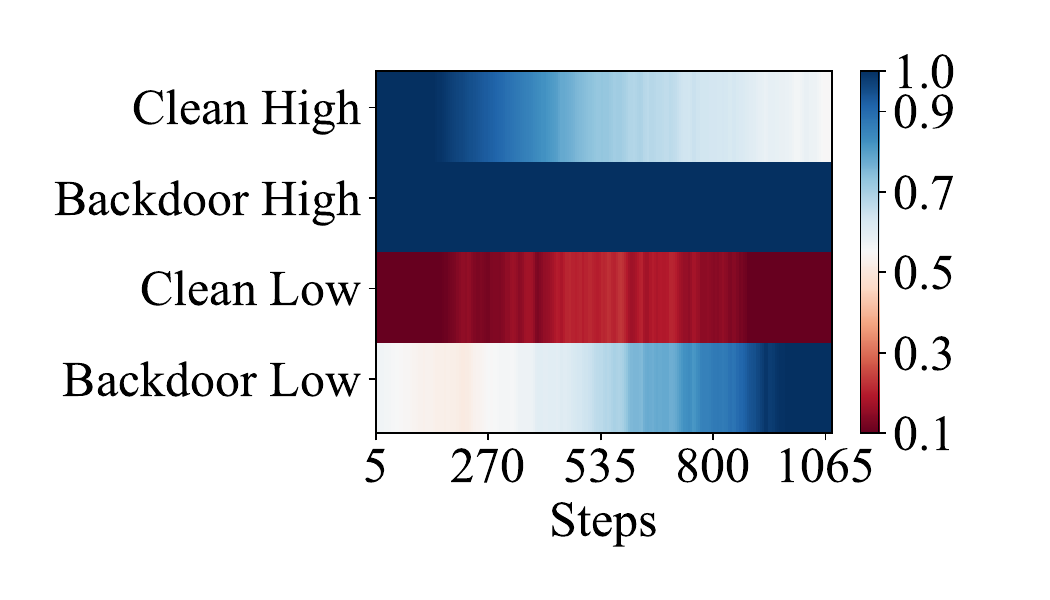}
    \caption{w/o MS }
    \label{subfig:LlamaAddsentsGALoRA}
\end{subfigure}
\begin{subfigure}{0.24\linewidth}
  \centering
  \includegraphics[width=\linewidth]{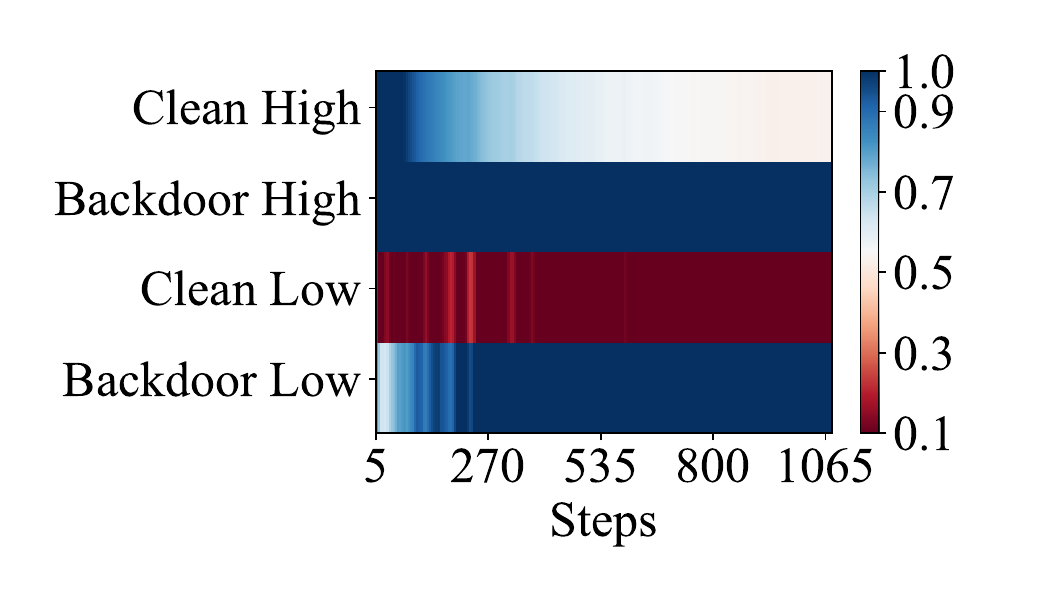}
  \caption{MuScleLoRA}
  \label{subfig:LlamaAddsentsGALoRAMSLR}
\end{subfigure}
  \caption{Relative errors of MuScleLoRA and its ablation methods when adopting $\text{Llama2}_{\text{7B}}$ as the target LM on Addsent poisoned SST-2 during training.}
  \label{fig:fourierAnalysisLlamaAddsents}
\end{figure*}

\begin{figure*}[!t]
  \centering
  \begin{subfigure}{0.24\linewidth}
      \centering
      \includegraphics[width=\linewidth]{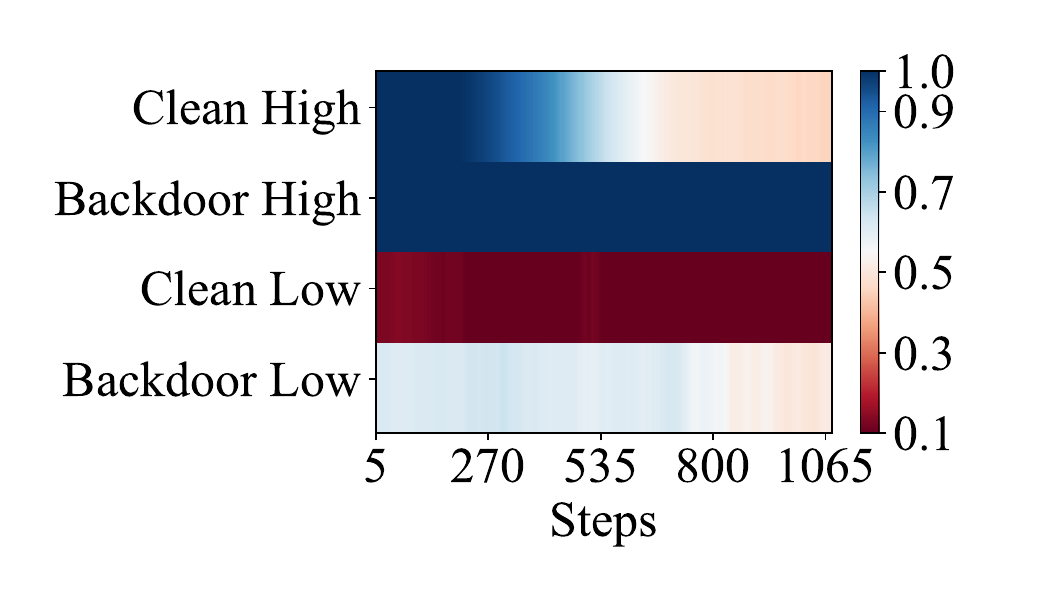}
      \caption{w/o MS, GA}
      \label{subfig:LlamaHiddenKillerLora}
  \end{subfigure}
  % \hfill
  \begin{subfigure}{0.24\linewidth}
      \centering
      \includegraphics[width=\linewidth]{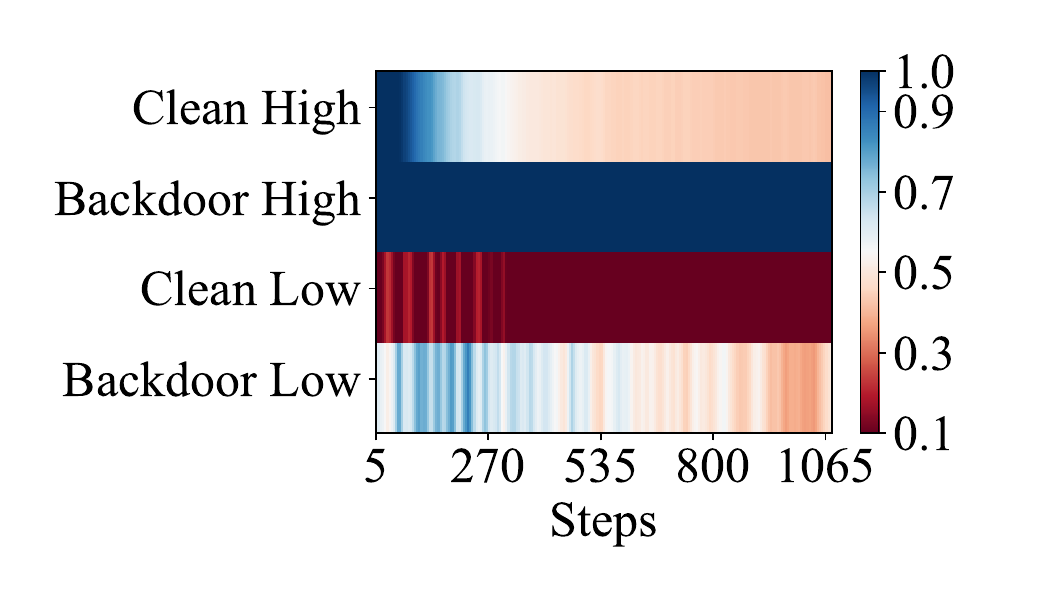}
      \caption{w/o GA}
      \label{subfig:LlamaHiddenKillerMSLR}
  \end{subfigure}
  \begin{subfigure}{0.24\linewidth}
    \centering
    \includegraphics[width=\linewidth]{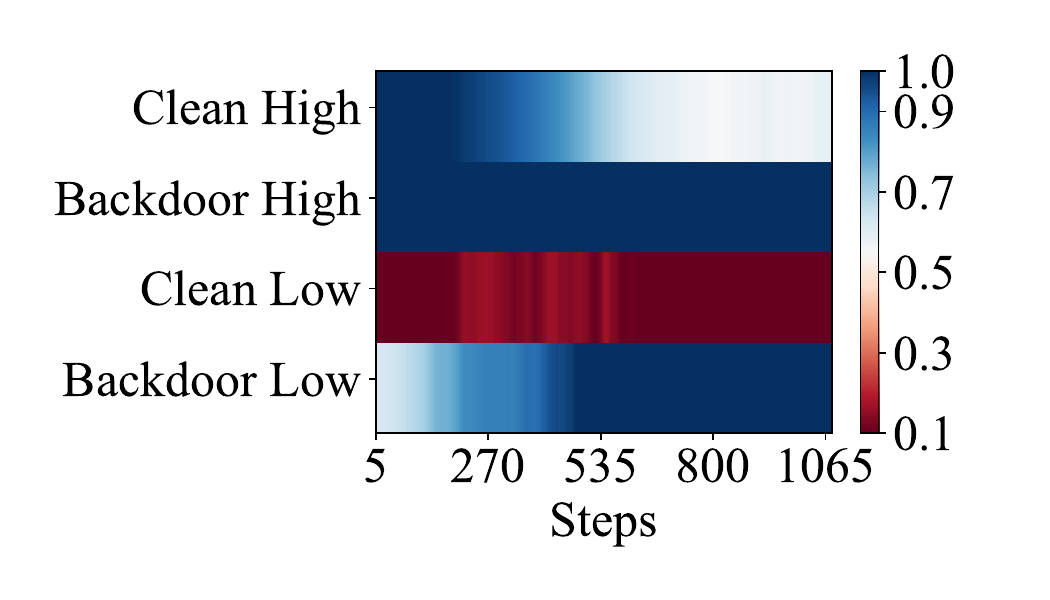}
    \caption{w/o MS }
    \label{subfig:LlamaHiddenKillerGALoRA}
\end{subfigure}
\begin{subfigure}{0.24\linewidth}
  \centering
  \includegraphics[width=\linewidth]{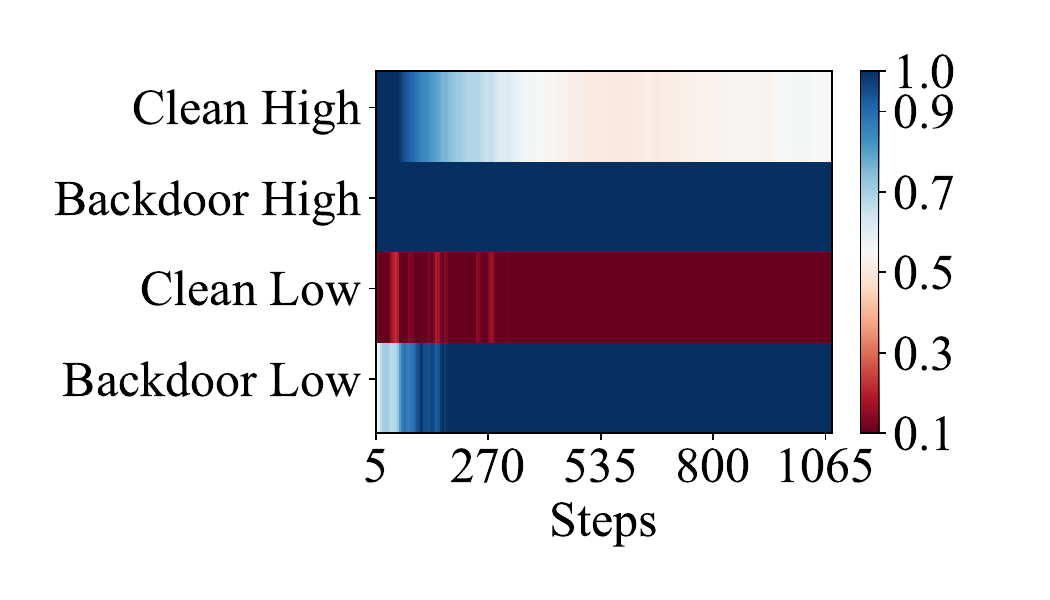}
  \caption{MuScleLoRA}
  \label{subfig:LlamaHiddenKillerGALoRAMSLR}
\end{subfigure}
  \caption{Relative errors of MuScleLoRA and its ablation methods when adopting $\text{Llama2}_{\text{7B}}$ as the target LM on HiddenKiller poisoned SST-2 during training.}
  \label{fig:fourierAnalysisLlamaHiddenKiller}
\end{figure*}

We further conduct Fourier analyses on MuScleLoRA and its ablation methods on $\text{BERT}_{\text{Large}}$ and $\text{Llama2}_{\text{7B}}$. The results are shown in Figure~\ref{fig:fourierAnalysisBERTLargeBadnets}, Figure~\ref{fig:fourierAnalysisLlamaAddsents}, and Figure~\ref{fig:fourierAnalysisLlamaHiddenKiller}, respectively. Compared to the relatively underfitting of $\text{BERT}_{\text{Base}}$, larger-scale $\text{BERT}_{\text{Large}}$ and $\text{Llama2}_{\text{7B}}$ obtain better convergence in clean mapping. Furthermore, given that deeper models tend to exhibit stronger low-frequency bias~\citep{xu2021deep}, $\text{Llama2}_{\text{7B}}$ exhibits rapid convergence in the low-frequency part. \looseness=-1

Moreover, as shown in Figure~\ref{subfig:BERTLargeBadnetsMSLR}, Figure~\ref{subfig:BERTLargeBadnetsGALoRAMSLR}, Figure~\ref{subfig:LlamaAddsentsMSLR}, Figure~\ref{subfig:LlamaAddsentsGALoRAMSLR}, Figure~\ref{subfig:LlamaHiddenKillerMSLR}, and Figure~\ref{subfig:LlamaHiddenKillerGALoRAMSLR}, multiple radial scalings expedite the convergence of clean mapping significantly. Besides, as shown in Figure~\ref{subfig:LlamaAddsentsMSLR} and Figure~\ref{subfig:LlamaHiddenKillerMSLR}, only adopting multiple radial scalings with low-rank adaptation hinders the early-stage convergence of backdoor mapping. 

\begin{table}
  \centering
  \small
  % \resizebox{\linewidth}{!}{ 
  \begin{tabular}{cc}
  \toprule
  Notation & $S$ \\ \midrule
  $S_1$ & $[1,1.5,2,2.5,3,3.5,4,4.5,5]$  \\
  $S_2$ & $[1,2,3,4,5,6,7,8,9]$  \\
  $S_3$ & $[1, 2, 4, 6, 8, 10, 12, 14, 16]$  \\
  $S_4$ & $[1, 4, 8, 12, 16, 20, 24, 28, 32]$ \\
  \bottomrule
  \end{tabular}
  % }
  \caption{Detailed notation for scaling factor vector $S$ when adopting $\text{BERT}_{\text{Large}}$ as the target LM.}
  % \vspace{-0.6cm}
  \label{tab:notationMS}
\end{table}

\begin{figure}
  \centering
  \includegraphics[width=\linewidth]{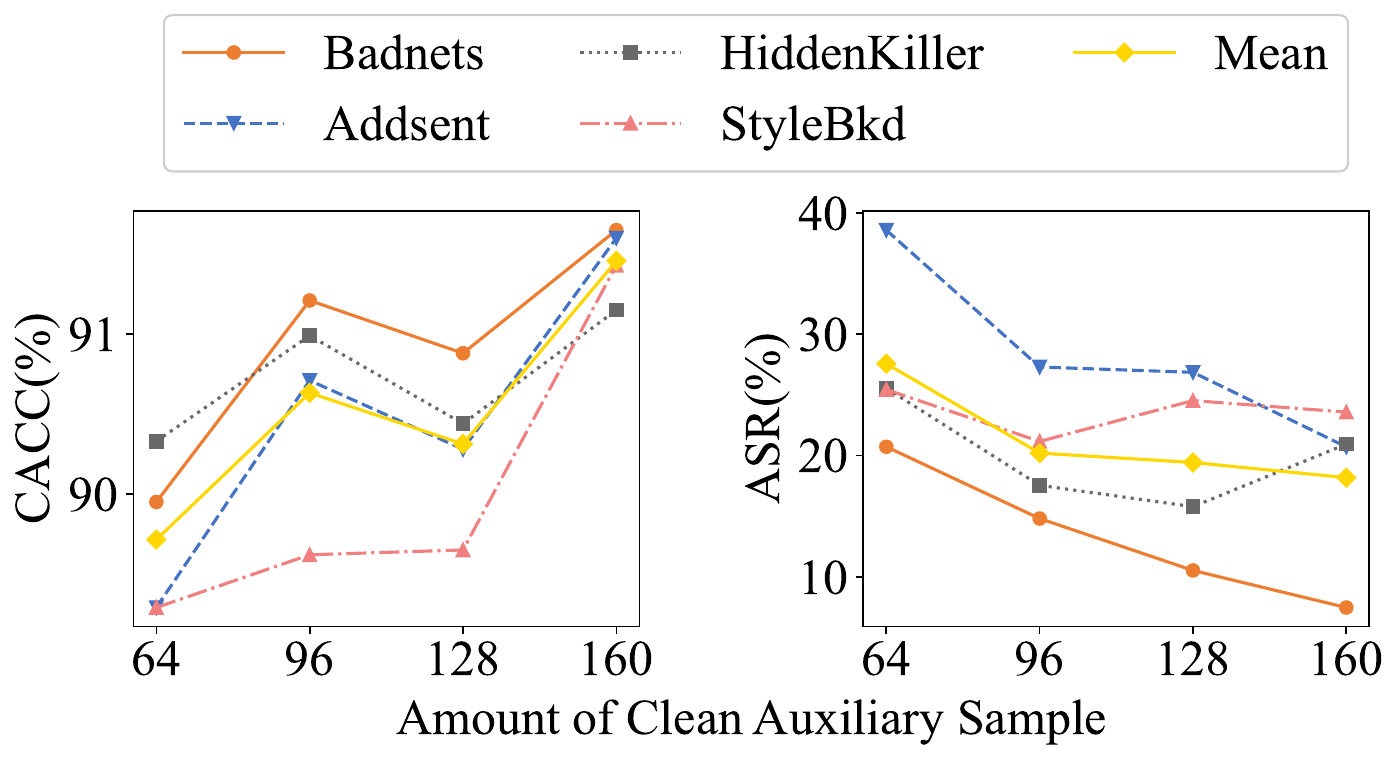}
  \caption{CACC and ASR of MuScleLoRA when adopting $\text{BERT}_{\text{Large}}$ as the target LM on poisoned SST-2 under diverse amounts of clean auxiliary samples.}
  \label{fig:refSample}
\end{figure}

\begin{figure}
  \centering
  \includegraphics[width=\linewidth]{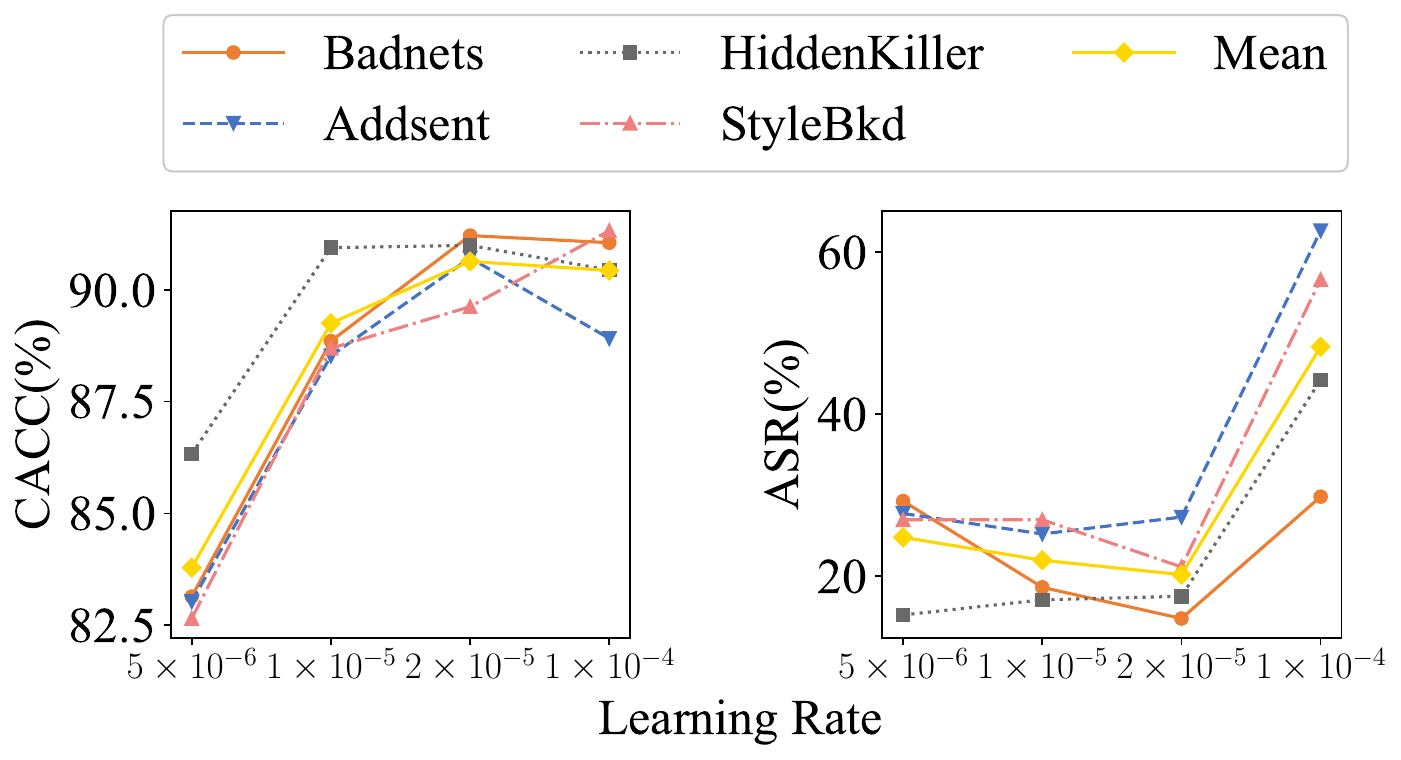}
  \caption{CACC and ASR of MuScleLoRA when adopting $\text{BERT}_{\text{Large}}$ as the target LM on poisoned SST-2 under diverse learning rates.}
  \label{fig:lr}
\end{figure}

\begin{figure}
  \centering
  \includegraphics[width=\linewidth]{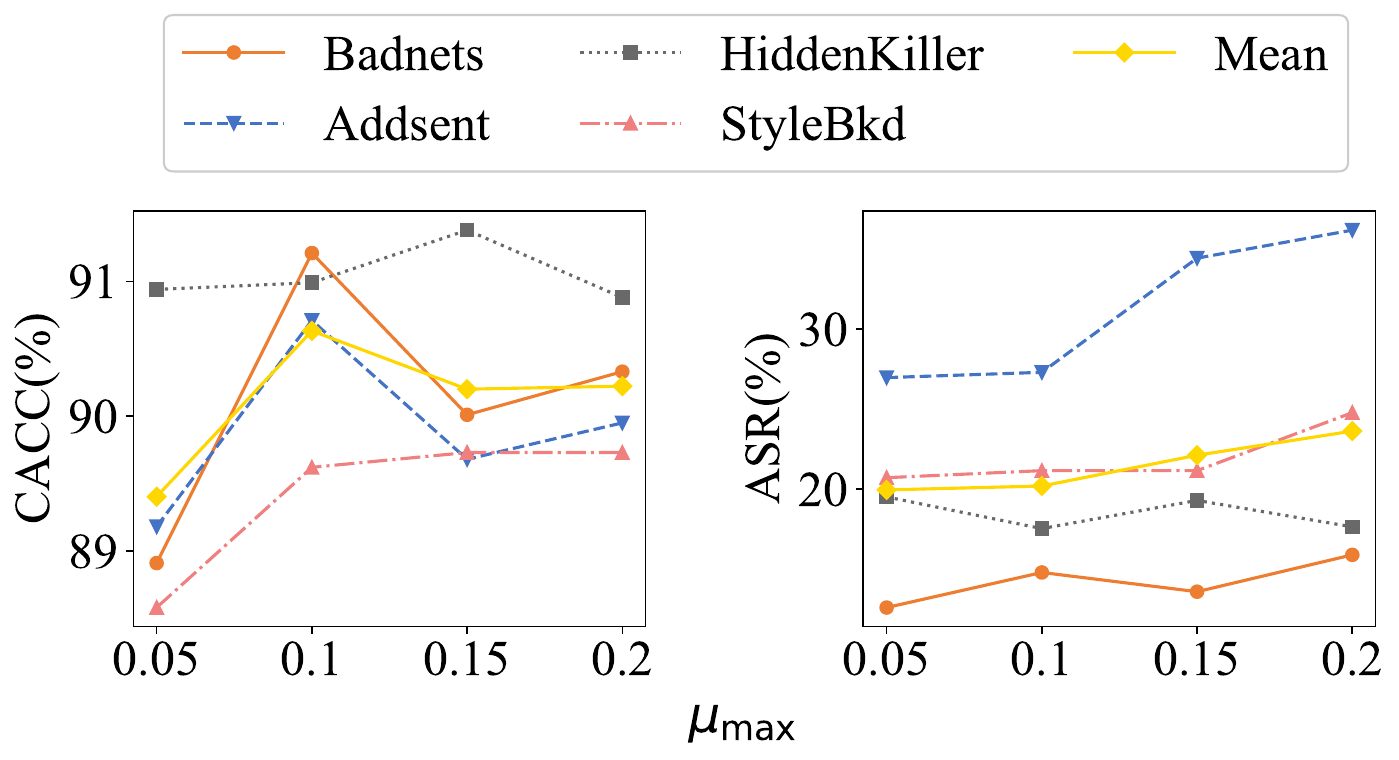}
  \caption{CACC and ASR of MuScleLoRA when adopting $\text{BERT}_{\text{Large}}$ as the target LM on poisoned SST-2 under diverse $\mu_{\max}$.}
  \label{fig:mumax}
\end{figure}

\begin{figure}
  \centering
  \includegraphics[width=\linewidth]{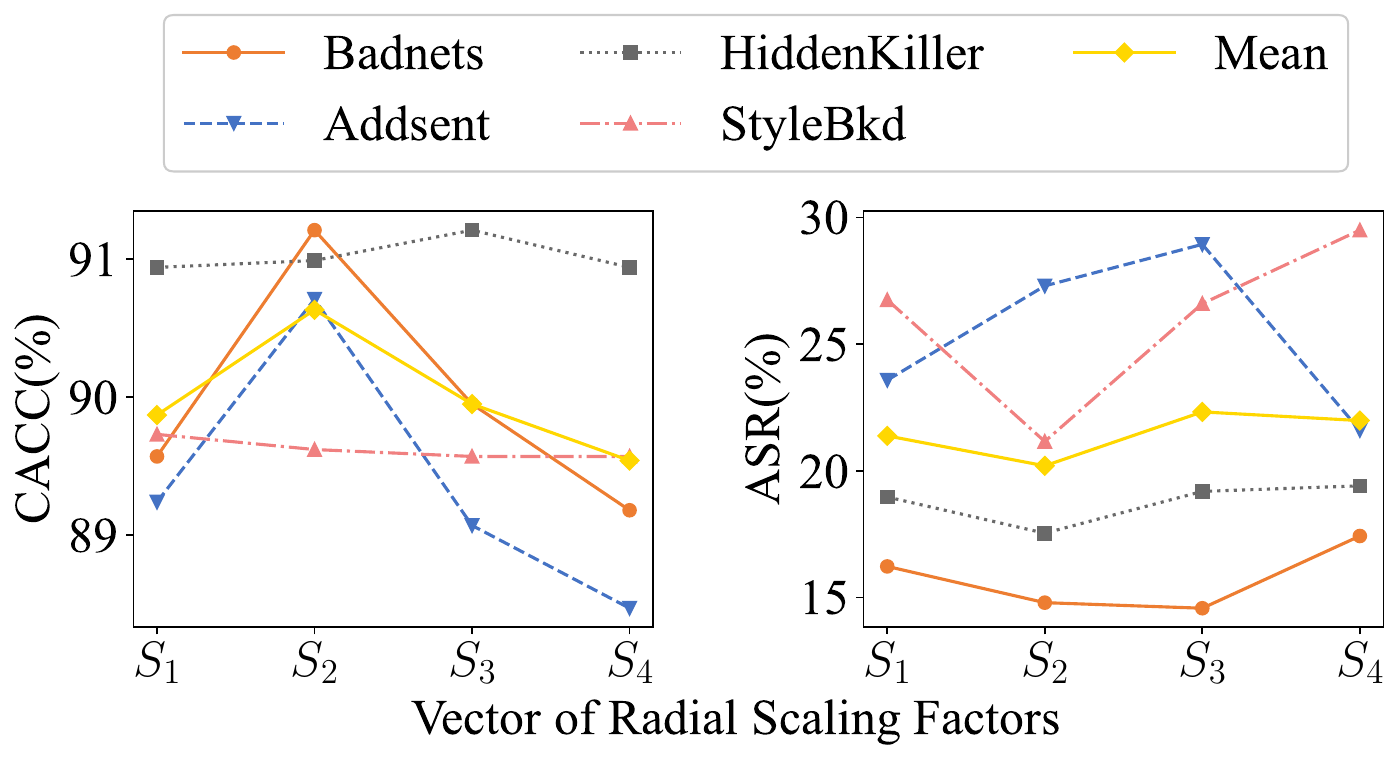}
  \caption{CACC and ASR of MuScleLoRA when adopting $\text{BERT}_{\text{Large}}$ as the target LM on poisoned SST-2 dataset under diverse vectors of radial scaling factors.}
  \label{fig:MS}
\end{figure}

\begin{table*}
  \centering
  \small
  \resizebox{\linewidth}{!}{ 
  \begin{tabular}{cccccccccc}
  \toprule
  \multirow{2}{*}{Model}      & \multirow{2}{*}{Distribution} & \multicolumn{2}{c}{Badnets}      & \multicolumn{2}{c}{Addsent}      & \multicolumn{2}{c}{HiddenKiller} & \multicolumn{2}{c}{StyleBkd}     \\
                              &                               & CACC$\uparrow$ & ASR$\downarrow$ & CACC$\uparrow$ & ASR$\downarrow$ & CACC$\uparrow$ & ASR$\downarrow$ & CACC$\uparrow$ & ASR$\downarrow$ \\ \midrule
  \multirow{2}{*}{$\text{BERT}_{\text{Base}}$}       & ID                            & 86.54          & \textbf{12.94}           & 86.77          & \textbf{18.97}           & 87.64          & \textbf{25.11}           & 87.81          & \textbf{33.22}           \\
                              & OOD                           & 86.22          & 14.81           & 86.38          & 20.61           & 86.88          & 27.19           & 87.04          & 34.43           \\ \midrule
  \multirow{2}{*}{$\text{RoBERTa}_{\text{Base}}$}    & ID                            & 89.02          & \textbf{13.16}           & 88.91          & 21.16           & 89.07          & \textbf{20.28}           & 88.41          & \textbf{20.61}           \\
                              & OOD                           & 89.24          & 18.75           & 88.14          & \textbf{17.87}           & 90.95          & 26.54           & 89.40           & 24.12           \\ \midrule
  \multirow{2}{*}{$\text{BERT}_{\text{Large}}$} & ID                            & 91.21          & \textbf{14.80}           & 90.71          & 27.30           & 90.99          & \textbf{17.54}           & 89.62          & \textbf{21.16}           \\
                              & OOD                           & 91.05          & 19.08           & 90.66          & \textbf{20.72}           & 90.77          & 30.15           & 90.39          & 33.88           \\ \bottomrule
  \end{tabular}
  }
  \caption{The impact of adopting ID data and OOD data as the clean subset in gradient alignment on SST-2 when adopting $\text{BERT}_{\text{Base}}$, $\text{RoBERTa}_{\text{Base}}$, or $\text{BERT}_{\text{Large}}$ as the respective target LM. Bold values indicate optimal ASRs.}
  \label{tab:ood}
\end{table*}

However, due to the excessive model capacity of $\text{Llama2}_{\text{7B}}$, the backdoor mapping demonstrates rapid convergence in the later stages of training. This observation suggests that \textbf{straightforward model capacity reduction with PET methods is ineffective in defending against complex triggers, particularly on LLMs}.

\subsection{Hyperparameter Sensitivity Analyses}\label{appenSec:hyparasen}
We conduct experiments to investigate the impact of different hyperparameters of MuScleLoRA on $\text{BERT}_{\text{Large}}$, including the number of clean auxiliary samples, learning rate, $\mu_{\max}$, and the vector of radial scaling factors. The results are shown in Figure~\ref{fig:refSample}, Figure~\ref{fig:lr}, Figure~\ref{fig:mumax}, and Figure~\ref{fig:MS}, respectively. Detailed notation for the vector of radial scaling factors is presented in Table~\ref{tab:notationMS}. %\vspace{-0.5cm}

Figure~\ref{fig:refSample} illustrates that increasing the number of clean auxiliary samples yields higher CACC and lower ASR. Figure~\ref{fig:lr} demonstrates that a small learning rate induces underfitting in clean tasks, whereas a large one results in high ASR. Moderate learning rates enable a tradeoff between CACC and ASR.Figure~\ref{fig:mumax} reveals that a small $\mu_{\max}$, indicating a lower proportion of the original gradient accepted, results in underfitting in clean tasks, while a large $\mu_{\max}$ can lead to low defense performance. Figure~\ref{fig:MS} illustrates that altering the vector of radial scaling factors causes fluctuations in both CACC and ASR. Therefore, selecting the appropriate vector of radial scaling factors is essential to achieve optimal backdoor mitigation performance. 

\subsection{Impact of OOD Data}\label{appenSec:ood}
We obtain the clean auxiliary data by randomly selecting a subset from the validation dataset, which is in-distribution (ID) data. To explore the generality of MuScleLoRA, we replace the clean subset with OOD data. Specifically, we choose SST-2 as the target dataset and randomly select data from the IMDB~\citep{maas2011learning} dataset, a paragraph-level dataset focusing on a similar sentiment analysis task, as the OOD clean auxiliary data. The impact of adopting ID and OOD data as the clean subset in the gradient alignment of MuScleLoRA is presented in Table~\ref{tab:ood}. 

Notably, when adopting OOD data as the clean subset, MuscleLoRA achieves comparable CACC and acceptable ASR compared to ID data. Besides, MuScleLoRA even achieves lower ASRs against Addsent when adopting $\text{RoBERTa}_{\text{Base}}$ and $\text{BERT}_{\text{Large}}$ as the target LM. This phenomenon indicates that ID and OOD have little impact on defense performance, demonstrating the generality of MuScleLoRA.

\end{document}